\documentclass[letterpaper, 10 pt, journal, twoside]{IEEEtran}
\ifCLASSINFOpdf
\else
\fi

\usepackage{subfiles}
\usepackage[dvipsnames]{xcolor}
\definecolor{myred}{rgb}{0.7, 0.0, 0.0}
\definecolor{mygreen}{rgb}{0.0, 0.7, 0.0}
\definecolor{myblue}{rgb}{0.0, 0.0, 0.8}

\usepackage{caption}
\usepackage{subcaption}
\usepackage{epsfig}
\usepackage{amsmath,bm}
\usepackage{amssymb} 
\usepackage{graphicx}
\usepackage{balance}

\newcommand{\rev}[1]{\textcolor{black}{#1}}

\usepackage{fancyhdr}
\usepackage{hyperref}

\hypersetup{
    colorlinks=true,
    linkcolor=black,
    citecolor=black,
    filecolor=black,
    urlcolor=black,
}

\begin{document}

\pagestyle{empty}
\begin{minipage}[b]{0.9\textwidth}
\begin{center}
This paper has been accepted for publication in IEEE Robotics and Automation Letters.
\end{center}
\vspace{0.5cm}
\begin{center}
DOI: \href{https://ieeexplore.ieee.org/document/9681286}{10.1109/LRA.2022.3142905} 
\end{center}
\begin{center}
IEEE Xplore: \href{https://ieeexplore.ieee.org/document/9681286}{https://ieeexplore.ieee.org/document/9681286}
\end{center}
\vspace{0.5cm}
\copyright 2022 IEEE.  Personal use of this material is permitted.  Permission from IEEE must be obtained for all other uses, in any current or future media, including reprinting/republishing this material for advertising or promotional purposes, creating new collective works, for resale or redistribution to servers or lists, or reuse of any copyrighted component of this work in other works.
\end{minipage}
\clearpage
%
\title{A Model for Multi-View Residual Covariances\\ based on Perspective Deformation}
%
%
%

\author{Alejandro Fontan, Laura Oliva, Javier Civera and Rudolph Triebel%
\thanks{Manuscript received September 9, 2021; accepted December 22, 2021. Date
of publication; date of current version. This letter was recommended for pub-
lication by Associate Editor Henrik Andreasson and Editor Sven Behnke upon
evaluation of the reviewers’ comments. This work was supported in part by
German Aerospace Center (DLR), in part by Spanish Government under Grant
PGC2018-096367-B-I00, and in part by Aragon Government under Grant DGA
T45 17R/FSE.}
\thanks{Alejandro Fontan is with the School of Engineering, University of Zaragoza,
50009 Zaragoza, Spain, and also with German Aerospace Center (DLR), Perception and Cognition Department, Institute of Robotics and Mechatronics, 82234
Weßling, Germany (e-mail: fontanvillacampa@gmail.com).} 
\thanks{Laura Oliva is with German Aerospace Center (DLR), Perception and Cog-
nition Department, Institute of Robotics and Mechatronics, 82234 Weßling,
Germany (e-mail: lauraolivamaza@gmail.com).}%
\thanks{Javier Civera is with the School of Engineering, University of Zaragoza, 50009
Zaragoza, Spain (e-mail: jcivera@unizar.es).}%
\thanks{Rudolph Triebel is the leader of the Department of Perception and Cognition Department, Institute of Robotics and Mechatronics, German Aerospace Center (DLR),  82234 Weßling, Germany. He is also affiliated as a guest professor at the Department of Aerospace and Geodesy, Technical University of Munich, 80333, München, Germany (e-mail rudolph.triebel@dlr.de).}%
\thanks{Digital Object Identifier 10.1109/LRA.2022.3142905}
}
%
%

\markboth{IEEE Robotics and Automation Letters. Preprint Version. Accepted December, 2021}
{Fontan \MakeLowercase{\textit{et al.}}: A Model for Multi-View Residual Covariances based on Perspective Deformation} 

%



\maketitle

\begin{abstract}
In this work, we derive a model for the covariance of the visual residuals in multi-view SfM, odometry and SLAM setups. The core of our approach is the formulation of the residual covariances as a combination of geometric and photometric noise sources. And our key novel contribution is the derivation of a term modelling how local 2D patches suffer from perspective deformation when imaging 3D surfaces around a point. Together, these add up to an efficient and general formulation which not only improves the accuracy of both feature-based and direct methods, but can also be used to estimate more accurate measures of the state entropy and hence better founded point visibility thresholds. We validate our model with synthetic and real data and integrate it into photometric and feature-based Bundle Adjustment, improving their accuracy with a negligible overhead.
\end{abstract}


%
\IEEEpeerreviewmaketitle

\section{Introduction}\label{sec:intro}
    
\begin{figure}
    \centering
    \includegraphics[width=0.34\textwidth]{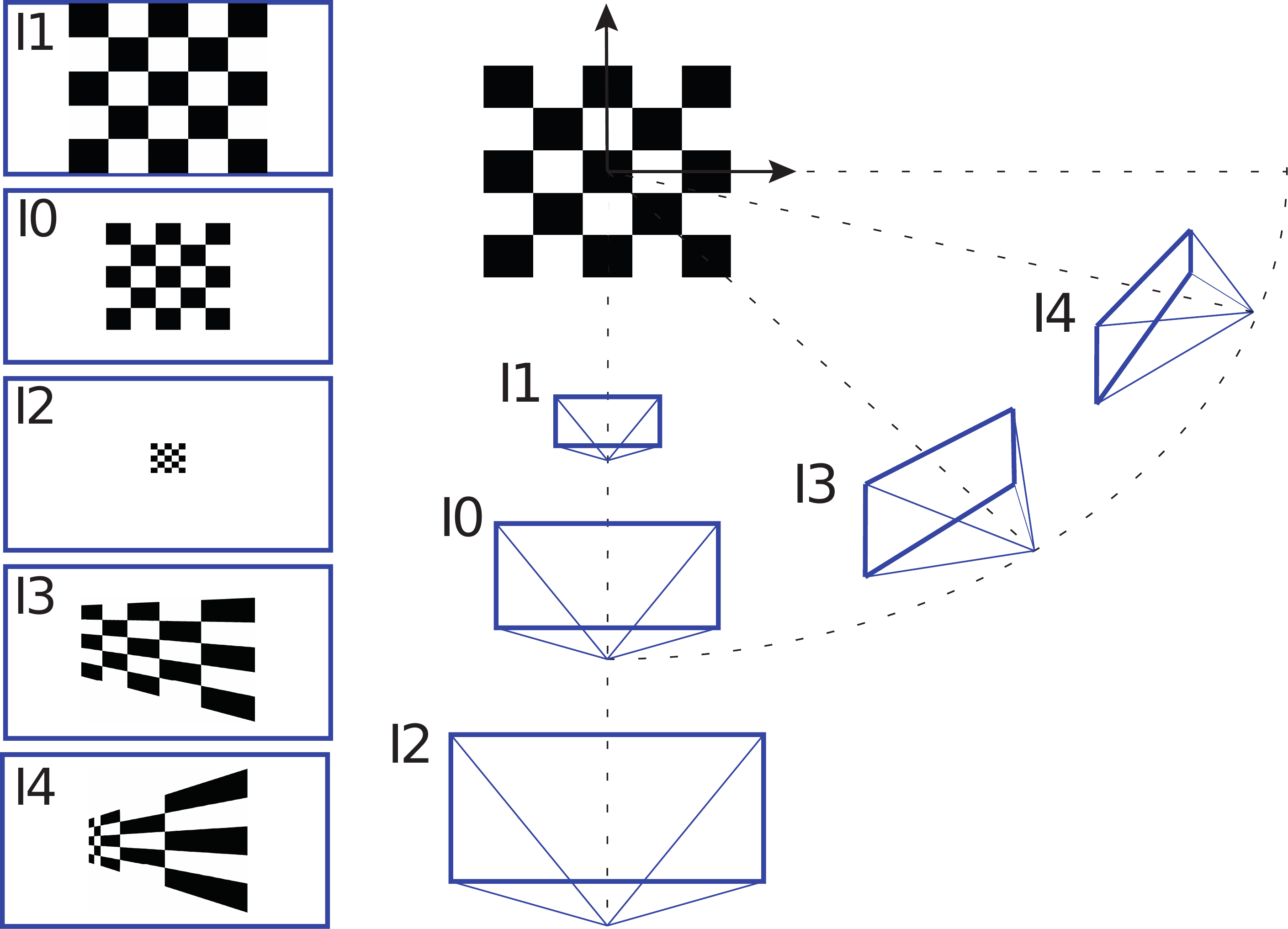}
    \includegraphics[width=0.09\textwidth]{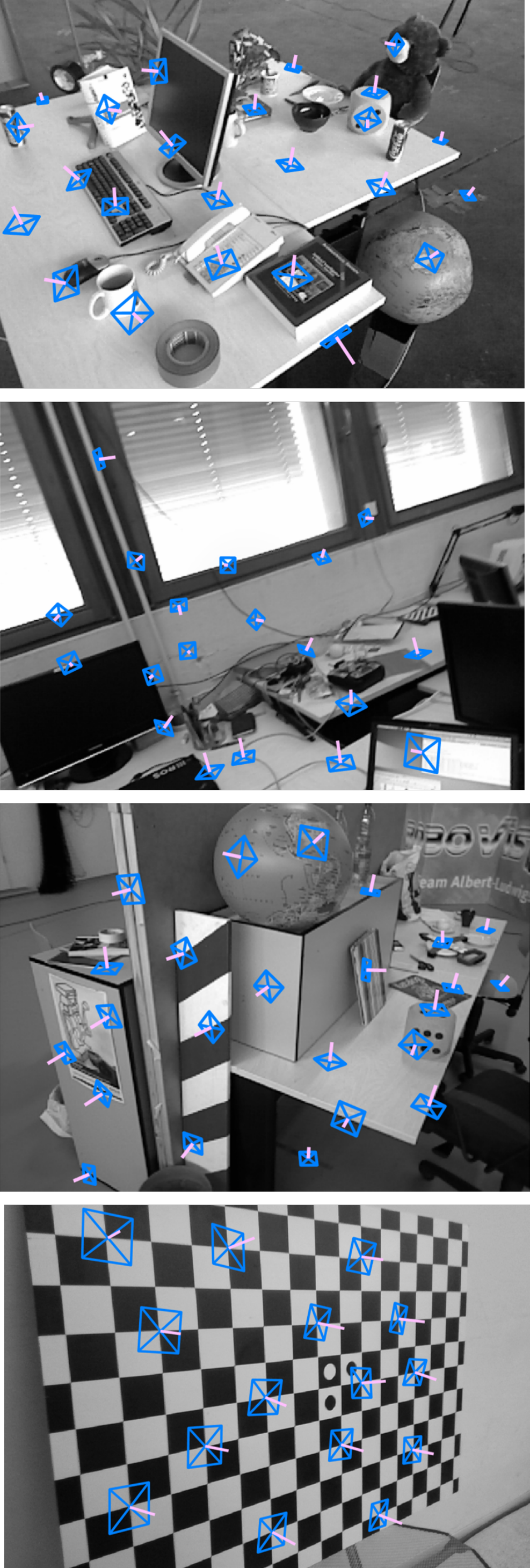}
    \caption{\textbf{Perspective deformation.} Image patches are subject to transformations when they are viewed from changing viewpoints, similar to the checkerboard in the image. This perspective deformation increments the covariance of the photometric/feature patches used by odometry and SLAM.}
    \label{fig:intro}
\end{figure}

\IEEEPARstart{W}{e} refer as perspective deformation to the transformations that apply to image patches when they are viewed from another viewpoint. Assuming constant camera intrinsics, it is the relative motion between a tridimensional surface and the camera poses what triggers perspective deformation in images. Figure \ref{fig:intro} shows an illustrative example where such deformations can be appreciated in a checkerboard pattern. In an abuse of language, throughout the paper we will use the terms traction and compression to characterize this perspective deformation. However, it should be remarked that we do not address deformable scenes but rigid environments.

Perspective deformation is a purely geometric effect and, yet, it is acknowledged as a challenge in many computer vision tasks. For example, SfM/odometry/SLAM pipelines, based on feature matching or photometric residuals, iterate over several pyramid levels \cite{engel2017direct} or set heuristic thresholds reflecting low confidence for wide-baseline matches \cite{mur2017orb}. The accuracy of a camera calibration can be modelled as the trade-off between a sufficiently informative geometric configuration and the image noise that perspective deformations produce \cite{peng2019calibration}. In other tasks such as semantic segmentation or object/place recognition, perspective deformation is also an issue if viewpoints vary significantly \cite{siddiqui2020viewal,hussain2016dealing,garg2019semantic}. 

For the specific case of multi-view reconstruction, when looking at Figure~\ref{fig:intro}, it is evident that a high degree of perspective deformations will also distort appearance-based descriptors, resulting in noisier image matches. However, visual residuals $\textbf{r}$ are modeled as isotropic Gaussians $\textbf{r} \sim \mathcal{N}\left( \boldsymbol{0}, \boldsymbol{\Sigma}_r\right), \boldsymbol{\Sigma}_r=\sigma_r^2 \textbf{I}$ in the vast majority of 3D vision pipelines. The visual residual model has a direct influence in the accuracy of the camera and structure states $\textbf{x}$ via the Gauss-Newton updates $\Delta \textbf{x} = - \left( \textbf{J}^\top \boldsymbol{\Sigma}_r \textbf{J} \right)^{-1}\textbf{J}^\top \textbf{r}$ ($\textbf{J}$ stands here for the derivatives of the residuals $\textbf{r}$ with respect to $\textbf{x}$). In this paper we propose a new model for the covariances of the visual residuals $\sigma_r^2$ that accounts for the effect of the perspective deformation and hence improves the accuracy of multi-view structure and motion estimations.

Furthermore, as \cite{peng2019calibration} points out, in many practical applications one should incorporate estimates of the uncertainty when available. The covariances $\boldsymbol{\Sigma}_x$ over the state $\textbf{x}$ are usually back-propagated from the residual covariance $\boldsymbol{\Sigma}_r$ as $\boldsymbol{\Sigma}_x = (\textbf{J}^T\boldsymbol{\Sigma}_r^{-1}\textbf{J})^{-1}$  \cite{strasdat2010real}. A better model for the residual covariances $\boldsymbol{\Sigma}_r$ would lead to more realistic uncertainty estimates, which is crucial in real-world applications.

\begin{figure}
\centering
\begin{subfigure}[b]{\columnwidth}
    \includegraphics[scale=0.2]{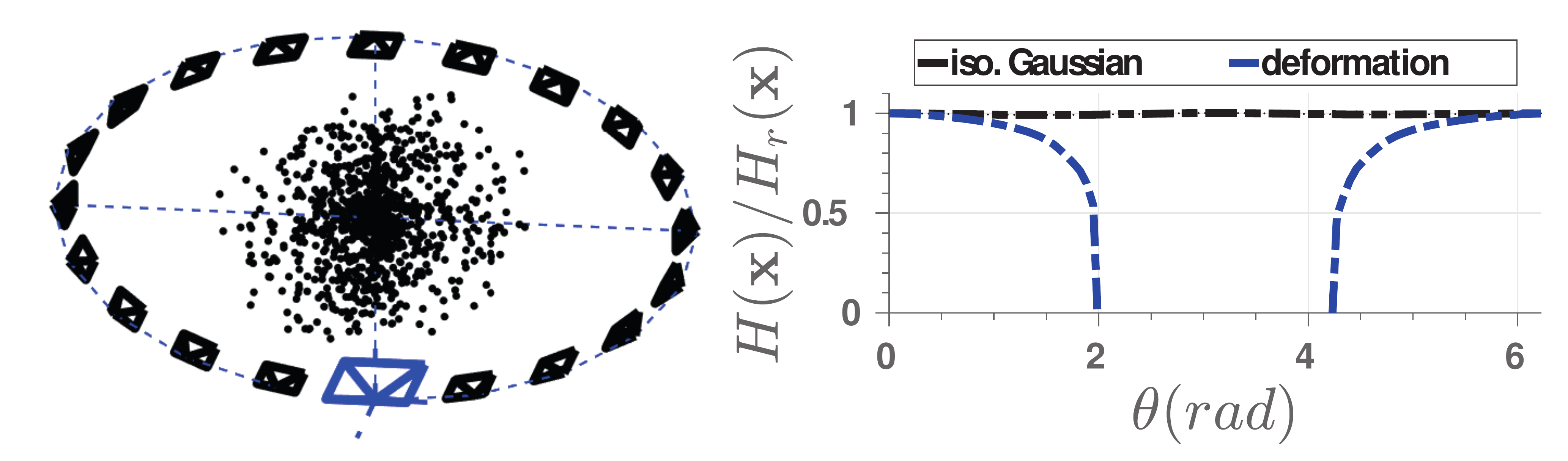}
    \caption{\textbf{Circular trajectory}. For isotropic Gaussian residuals, the differential entropy is incorrectly modeled as constant even for $180^\circ$ parallax. Our deformation-based covariance models it correctly, showing a steep decrease with parallax.}
\end{subfigure}
\begin{subfigure}[b]{\columnwidth}
    \includegraphics[scale=0.2]{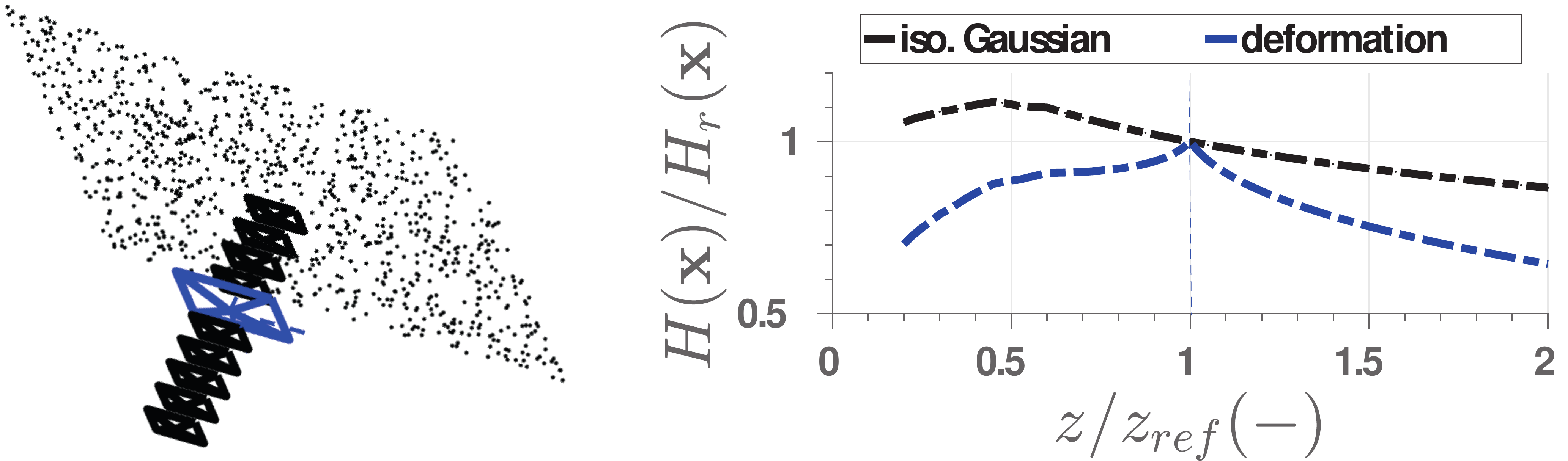}
    \caption{\textbf{Approaching trajectory}. For isotropic Gaussian residuals, the differential entropy increases as the camera approaches the planar scene, which is not correct. With our deformation-based model areas close to the reference frame (the blue one) give the best accuracy.}
\end{subfigure}
\caption{Differential entropy inconsistencies. A deficient model of the residuals leads to inconsistencies in a variety of applications. An illustrative one is the camera pose entropy in these two situations.}
\label{fig:entropyInconsistencies}
\end{figure}

As a final application case, using information metrics in odometry and SLAM dates back to works such as \cite{strasdat2010real,chli2008active,kerl2013dense} but has seen great progress recently ---aiming to the reduction of the computational demand for their implementation on low-end platforms \cite{fontan2020information,kuo2020redesigning, zhao2018good,zhao2020good}. Again, a better model for $\boldsymbol{\Sigma}_r$ would significantly improve such approaches. As two illustrative examples, Figure \ref{fig:entropyInconsistencies} shows inconsistencies that arise when the differential entropy $H(x) = -\frac{1}{2}\log((2\pi e)^k|\boldsymbol{\Sigma}_x|)$ is obtained approximating the residual distribution by an isotropic Gaussian.

As a summary, the specific contributions of this paper are as follows. First, we derive a model for the perspective deformation of 2D image patches (Section \ref{sec:perspDef}). To the best of our knowledge, we are the first ones addressing such deformation in a general manner. Second, we introduce a model for the visual residuals based on the perspective deformation, valid for both feature-based and photometric methods (Section \ref{sec:visualCov}). Third, we validate our model with extensive experimentation in a realistic synthetic dataset and real data (Section \ref{sec:modelVal}). To our knowledge, this is the first time that the relation between perspective deformation and multi-view residuals is shown and characterized in several setups under a unified derivation. Finally, we integrate our model in the global optimization of feature-based and direct odometry/SLAM pipelines, demonstrating a consistent reduction of the trajectory error in the TUM RGB-D dataset \cite{sturm12iros} (Section \ref{sec:experiments}).

\section{RELATED WORK} \label{sec:relatedWork}

Isotropic visual residuals are widespread in multi-view setups \cite{klein2007parallel,endres20133,wilson2014robust,schonberger2016structure,mur2017orb,engel2017direct,Rosinol20icra-Kimera} and only a few exceptions difer or are directly related to our work. The work of \cite{yang2018challenges} implicitly underlines the importance of visual covariances by evaluating aspects such as photometric calibration, motion bias and rolling shutter, and their effect on direct, feature-based, and semi-direct odometries.

Molton et al.\ \cite{molton2004locally} models salient features as observations of locally planar regions, compensating for the predicted motion before matching. Such early model is, however, limited to template matching based on cross-correlation. More recently, for the application of camera calibration, Peng and Sturm \cite{peng2019calibration} incorporate uncertainty for the corners of a calibration target using autocorrelation matrices.

Engel et al.\ \cite{engel2017direct} apply a gradient-dependent weighting, reducing the effect of photometric errors in pixels with high gradient. This can be probabilistically explained as approximating the geometric error by adding on the projected point position, small and independent geometric noise, and directly marginalizing it. Mur-Artal and Tard{\'o}s \cite{mur2017orb} scale the visual residual proportionally to the resolution where the ORB features are detected. In both works, apart from these two aspects, the noise model follows the standard isotropic Gaussian assumption. Up to our knowledge, ours is the first model deriving a probabilistic form of perspective deformation, opening a research line towards a better understanding and a general modeling of visual residuals.

\section{PERSPECTIVE DEFORMATION} \label{sec:perspDef}
    \subsection{Preliminaries} \label{subsec:preliminaries}

We refer with subscript $j$ to the reference frame where a 3D point $\textbf{p} \in \mathbb{R}^3$ is first observed, and with $i$ to any other frame from which the point is visible. The image coordinates of the projection of \textbf{p} in reference frame $j$ and its depth are denoted as $\textbf{u} \in \boldsymbol{\Omega}$ and $z \in \mathbb{R}$ respectively, where $\boldsymbol{\Omega}$ is the image domain.

The function $\varphi(\textbf{u})$ projects a point $\textbf{p}$ from its camera coordinates $\textbf{u}$ in the reference frame $j$ into the frame $i$,

\begin{equation}
\varphi(\textbf{u}) = \Pi(\textbf{R}\Pi^{-1}(\textbf{u},z) + \textbf{t}),
\label{eq:projFunct}
\end{equation}

\noindent where $\Pi(\textbf{p})$ (determined by the intrinsic camera parameters) projects the point $\bf{p}$ in the camera frame; and $\Pi^{-1}(\textbf{u},z)$ back-projects the image point with coordinates $\bf{u}$ at depth $z$. $\textbf{R} \in \mathrm{SO}(3)$ and $\textbf{t} \in \mathbb{R}^3$ are the relative rotation and translation between frame $j$ and frame $i$.

    \subsection{Surface representation}  \label{subsec:surface}

We consider that each point $\textbf{p}$ lays on a local 3D surface $S$. Our formulation can include surfaces with any degree of complexity as far as the depth $z$ of $\textbf{p}$ can be expressed as a function of its image coordinates $z = S(\textbf{u})$. Similarly to \cite{molton2004locally}, in our implementation we constrain those surfaces to be 3D planes $S = f(\alpha,\beta, \gamma)$; with $\alpha$, $\beta$ and $\gamma$ being the plane parameters on the camera reference frame. We then can formulate the depth $z$ for each point $\textbf{p}$ in terms of its camera coordinates $\textbf{u}$ and its corresponding local plane parameters $(\alpha, \beta, \gamma)$:

\begin{equation}
z = \frac{\gamma}{1 - [\alpha,\beta,0 ]\cdot \Pi^{-1}(\textbf{u},1)} .
\label{eq:surface}
\end{equation}

Commonly, direct VO and SLAM extend point descriptors over a small neighborhood of pixels \cite{engel2017direct}. Feature-based pipelines \cite{schonberger2016structure,mur2017orb} with classic feature descriptors \cite{lowe2004distinctive,6126544} also perform operations on patterns around a central pixel. Operating on a pattern of pixels on the image is equivalent to consider that all these pixels have the same local depth coordinate ($\alpha = 0 , \beta = 0, \gamma = z$). Note that our former assumption is a compromise between considering more complex surfaces and assuming that the point belongs to a plane orthogonal to the local z-axis.

    \subsection{Perspective deformation model}  \label{subsec:perspDef}

We will approximate the projection function of Equation \eqref{eq:projFunct} by its \textbf{first-order Taylor approximation}

\begin{equation}
\varphi(\textbf{u}+ d\textbf{u}) \approx \varphi(\textbf{u}) + \nabla_{\textbf{u}} \varphi d\textbf{u} .
\label{eq:linearization}
\end{equation}

The \textbf{perspective deformation gradient tensor} \rev{$\textbf{F}_{\psi}(\textbf{u})$} contains all the information about the  \rev{local rotation $\psi$} and deformation of $\textbf{u}$ and corresponds to the Jacobian matrix of the transformation $\varphi(\textbf{u})$,

\begin{multline}
\textbf{F}_\psi(\textbf{u}) = \nabla_{\textbf{u}} \varphi = \frac{\partial\varphi(\textbf{u})}{\partial \textbf{u}} =  \\ \Bigg[ \frac{\partial \textbf{u}}{\partial \textbf{p}_n}\frac{\partial \textbf{p}_n}{\partial \textbf{p}_c}\frac{\partial \textbf{p}_c}{\partial \textbf{p}_w}\Bigg]_i\Bigg[\frac{\partial \textbf{p}_w}{\partial \textbf{p}_c}\frac{\partial \textbf{p}_c}{\partial \textbf{p}_n}\frac{\partial \textbf{p}_n}{\partial \textbf{u}}\Bigg]_j ,
\label{eq:defGradTensor}
\end{multline}

\noindent where subscripts $n$, $c$ and $w$ correspond to the point coordinates normalized, in the camera frame and in the absolute one respectively. Note that the gradient tensor \rev{$\textbf{F}_\psi(\textbf{u})$} models how an infinitesimal line segment in the ``undeformed'' reference frame is not only stretched but also \rev{rotated with an angle $\psi$} into a line segment in the ``deformed'' frame.

The \textbf{Cauchy–Green deformation tensor} gives a measure of the deformation that is independent of the rotation around the camera axis, without needing explicitly the rotation matrix. By applying the polar decomposition theorem, which states that any second-order tensor can be decomposed into a product of a pure rotation and symmetric tensor, it is possible to separate the camera rotation $\textbf{R}_\psi$ from \rev{a rotation-independent deformation gradient tensor $\textbf{F}_u$, hence $\textbf{F}_\psi(\textbf{u}) = \textbf{R}_\psi \textbf{F}_u = \bar{\textbf{F}}_u\textbf{R}_\psi$}.

The tensor $\textbf{C}$ is called the right Cauchy-Green deformation tensor
\rev{
\begin{equation}
\textbf{C} = \textbf{F}_{\psi}(\textbf{u})^T\textbf{F}_\psi(\textbf{u})  = \textbf{F}_u^T \textbf{R}_\psi^T \textbf{R}_\psi \textbf{F}_u = \textbf{F}_u^T\textbf{F}_u .
\end{equation}
}
Since it is formed only from the $\textbf{F}_u$ tensor, it describes the deformation of the material ``before" rotation. 
 
The left Cauchy–Green deformation tensor $\bar{\textbf{C}}$
\rev{
 \begin{equation}
\bar{\textbf{C}} = \textbf{F}_\psi(\textbf{u})\textbf{F}_\psi(\textbf{u})^T  = \bar{\textbf{F}}_u\textbf{R}_\psi \textbf{R}_\psi^T  \bar{\textbf{F}}_u^T = \bar{\textbf{F}}_u\bar{\textbf{F}}_u^T ,
\label{eq::leftCauchyTensor}
\end{equation}
}
\noindent applies a rigid body rotation first, and then deforms the rotated volume. Both tensors are independent of the rotation, but they describe the deformation in different frames.

\textbf{Deformation.} Physically, the Cauchy–Green tensor gives us the square of the local geometric changes $\boldsymbol{\varepsilon}^2(\boldsymbol{\eta})$ due to deformation in some particular directions $\boldsymbol{\eta}$:

\begin{equation}
\boldsymbol{\varepsilon}^2(\boldsymbol{\eta}) =\boldsymbol{\eta}^T\textbf{C}(\textbf{u})\boldsymbol{\eta}.
\label{eq::perspDef}
\end{equation}

If we consider a single direction of interest $\boldsymbol{\eta} \in \mathbb{R}^2$ in which we want to obtain the perspective deformation (\emph{e.g.}, the direction of the gradient for photometric errors), we then obtain a scalar deformation $\varepsilon^2 \in \mathbb{R}$. On the other hand, if there are two directions of interest $\boldsymbol{\eta} \in \mathbb{R}^{2 \times 2}$ (\emph{e.g.}, the geometric residuals in feature-based methods) the deformation obtained is not only 2-dimensional but also anisotropic $\boldsymbol{\varepsilon}^2 \in \mathbb{R}^{2 \times 2}$.

\textbf{Traction and Compression} are relative terms that depend on which of the configurations we consider as ``undeformed''. Due to the linearized projection model (equations \eqref{eq:linearization} and \eqref{eq:defGradTensor}), we can derive that the inverse transformation $\textbf{F}^{-1}(\textbf{u})$ yields the inverse stretch $|\textbf{F}^{-1}(\textbf{u})| =|\textbf{F}(\textbf{u})|^{-1}$. In other words, we can map every compression $\varepsilon_c^2 \in [0,1)$ into its homologous traction $\varepsilon_t^2 \in (1,\infty)$ and viceversa just by $\varepsilon^2_t \approx \varepsilon^{-2}_c$.

\section{VISUAL RESIDUAL COVARIANCES} \label{sec:visualCov}

\textbf{Deformation Covariance.} Under traction ($\varepsilon^2 > 1$), the covariances of the visual residuals grow as a function of the deformation according to certain response functions $\sigma_{t}^2:\varepsilon^2 \subset \mathbb{R} \mapsto \mathbb{R}$. These functions vary with the residual model used for each particular application (e.g., feature-based or photometric) and we determine them experimentally in this work (see the validation in Section \ref{sec:modelVal} and further experiments in Section \ref{sec:experiments}). As mentioned, we will model an equivalent traction for every scalar compression with the response function $\sigma_{c}^2:\varepsilon^{-2} \subset \mathbb{R} \mapsto \mathbb{R}$. To sum up, for the case $\boldsymbol{\eta} \in \mathbb{R}^2$ our model results in

\begin{equation}
\sigma_\varepsilon^2(\varepsilon^{2}) = \left\{
\begin{array}{c l}
\sigma_c^2(\varepsilon^{-2}-1)& \varepsilon^2 \leq 1\\
  & \\
\sigma_t^2(\varepsilon^2-1) & \varepsilon^2 > 1 .
\end{array}
\right.
\label{eq:geoCovSplit}
\end{equation}

\noindent With this formulation we aim for compression and traction to have similar response functions that map deformations into visual covariances ($\sigma_t^2 = \sigma_c^2$). However, effects such as pixel discretization or processing done by feature-based approaches induce more complex behaviours for these response functions. We propose and analyze some particular cases in our validation experiments in Section \ref{sec:modelVal}.

\textbf{2d-deformation}. If residual covariances are coupled in two image directions $\boldsymbol{\sigma}^2_\varepsilon \in \mathbb{R}^{2\times2}$ (such as in corner matching), \rev{$\textbf{C} \in \mathbb{R}^{2 \times 2} $} is a diagonalizable symmetric positive semi-definite matrix. Then, it can be found a unitary matrix $\textbf{V} \in \mathbb{R}^{2 \times 2}$ \rev{where the matrix containing the deformation \eqref{eq::perspDef} in each direction $\boldsymbol{\varepsilon}^2_{kk} = \textbf{V}\textbf{C} \textbf{V}^T \in \text{diag}(\mathbb{R}^{2})$} is diagonal. We obtain the  non-diagonal covariance matrix applying the model in Equation \eqref{eq:geoCovSplit} to the diagonal elements of \rev{$\boldsymbol{\varepsilon}^2_{kk}$} and undoing the transformation \rev{$\boldsymbol{\sigma}^2_\varepsilon = \textbf{V}\sigma_\varepsilon^2(\boldsymbol{\varepsilon}^2_{kk}) \textbf{V}^T \in \mathbb{R}^{2 \times 2}$.}

\textbf{Projection Covariance}. In addition to perspective deformation, there are other possible noise sources (\emph{e.g.}, rolling shutter effects \cite{Choi_2015_CVPR,yang2018challenges}) that are propagated through the projection function to the visual residuals and can be added as geometric uncertainties in our covariance $\boldsymbol{\sigma}_\varphi^2$.

As an example, the depth uncertainty from stereo cameras and RGB-D ones using structured light can be propagated from the disparity variance $\sigma_\nu^2$. Assuming a focal length $f$, a baseline $b$ and a disparity $\nu$, the first-order propagation for the inverse depth covariance is \cite{handa2014benchmark,khoshelham2011accuracy,barron2013intrinsic,concha2017rgbdtam}

\begin{equation}
z = \frac{fb}{\nu}, \quad \sigma_z = \frac{fb}{\nu^2}\sigma_\nu = \frac{z^2}{fb}\sigma_\nu .
\label{eq:StereoCov}
\end{equation}

Using a first-order propagation of the projection in Equation \eqref{eq:projFunct}, we obtain the contribution of the depth uncertainty to the residual
\begin{equation}
\boldsymbol{\sigma}^2(z) = \bigg(\frac{\partial \textbf{u}}{\partial z}\bigg)^2\sigma_z^2 .
\label{eq:depthCov}
\end{equation}

Finally, all uncertainty contributions can be grouped together into a single term that models the full covariance of a visual measure in a given direction 

\begin{equation}
\boldsymbol{\sigma}_{\varphi}^2 =\boldsymbol{\sigma}_\varepsilon^2(\boldsymbol{\eta}) + \boldsymbol{\eta}^T(\boldsymbol{\sigma}^2(z) + ...)\boldsymbol{\eta} .
\label{eq:geoTerm}
\end{equation}

    \subsection{Implementation Details} \label{subsec:impDet}

For the sake of reproducibility, we describe several practical aspects of the implementation of our model, namely point visibility, photometric errors and feature matching.

\textbf{Photometric residual.} Direct methods define a photometric error between the raw image intensities $I:\boldsymbol{\Omega} \subset \mathbb{R}^2 \mapsto \mathbb{R}$. Although each method has specific particularities in their residual definitions, most of them concur in  evaluating the photometric error $r \in \mathbb{R}$ in a slightly spread pattern of pixels $\mathcal{P}$ \cite{engel2017direct,forster2014svo}

\rev{
\begin{equation}
       r = \sum_{u \in \mathcal{P}} (I_j(\textbf{u})-I_i(\varphi(\textbf{u})))^2 .
       \label{eq:photoErrorPattern}
\end{equation}
}

The residual covariance $\sigma_{r}^2$ can be modeled in this case as a purely photometric addend $\sigma_{N}^2$ and the geometric covariance from Equation \eqref{eq:geoTerm} \rev{propagated with the intensity gradient $G$}

\begin{equation}
\sigma_{r}^2 = \sigma_{N}^2 + G^2\sigma_{\varphi}^2(\boldsymbol{\eta}_g). 
\label{eq:photoCov}
\end{equation}

As shown in \cite{engel2016photometrically,engel2017direct,yang2018challenges}, a conscientious photometric calibration improves the accuracy and robustness of direct methods. It would be reasonable to include the model of the photometric contribution $\sigma_{N}^2$ at this point in the formulation. However, due to the scope of this paper, we include the propagation of the image noise obtained through bilinear interpolation $\sigma_{I}^2$ to the photometric error calculated on a N-sized pattern \eqref{eq:photoErrorPattern}

\begin{equation}
\sigma_N^2 = \frac{128N}{81}(\sigma_{I}^2)^2 .
\label{eq:patchNoise}
\end{equation}

If depth information is available, the pixels $\textbf{u} \in \mathcal{P}$ could be considered to belong to the same 3D surface (similar to Equation \eqref{eq:surface}) and reproject them accordingly. However, monocular setups estimate depth from multiple views, making the assumption that all pixels on a pattern share the same depth quite convenient. Note how our formulation can easily incorporate  this assumption with $\textbf{u} \in S_0 (\alpha = 0, \beta = 0, \gamma = z)$.

\textbf{Feature-based errors} are usually defined as variations of the following expression:
\begin{equation} \label{eq:geoCostFunction}
        \textbf{r} = \textbf{u}_i-\varphi(\textbf{u}),
\end{equation} 
where $\textbf{u}_i$ stands for the feature point in image $i$ and $\varphi(\textbf{u})$ for the corresponding point in the frame $j$ reprojected in the frame $i$. Hence the residual covariance is expressed as the sum of the projection covariance and the feature subpixel noise $\boldsymbol{\sigma}_u^2$:

\begin{equation} \label{eq:phCostFunction}
        \boldsymbol{\sigma}_r^2 =  \boldsymbol{\sigma}_u^2 +  \boldsymbol{\sigma}_\varphi^2
\end{equation}

\textbf{Right/left Cauchy-Green deformation tensor.} The right tensor $\textbf{C}$ models the deformation ``before" rotation, that means in the original frame. Since direct methods compute photometric gradients in these reference frame, it is not necessary to recompute them again. On the other hand, feature-based approaches set geometric residuals in the reprojection frame, that means ``after" rotation. Then by using the left Cauchy-Green tensor $\bar{\textbf{C}}$ deformations are conveniently referred to that coordinate frame.

\textbf{Point visibility.} Heuristic thresholds for point visibility are a trade-off between the potential benefits of wide baselines and the increasing matching uncertainties. Some approaches keep observations in a bunch of close keyframes and remove outliers with robust cost norms \cite{engel2017direct}. Others define angle and scale thresholds between viewing rays \cite{mur2017orb}. 
Defining a threshold in terms of deformation is more principled, since it accounts for the relative orientation between the camera and the local surface. If a point is observed with a parallax angle bigger than $90^\circ$, the diagonal of the deformation tensor \eqref{eq:defGradTensor} triggers a negative value. In addition, weighting visual residuals in terms of the perspective deformation allows a wider range of inliers without degrading the optimization.

\section{MODEL VALIDATION} \label{sec:modelVal}

First, we test the basis of our model equations (Section \ref{sec:perspDef}) with a Monte Carlo-based experiment. Next, to identify the applicable cases of the perspective deformation, and explore the function-revealing parameters of Equation \eqref{eq:geoCovSplit} that could guide good visual covariance modeling, we validate our approach in a two-branch experiment: photometric and feature-based. The estimates of the model parameters are then used as the input to the experiments in Section \ref{sec:experiments}. Finally,  we show the differences of taking into account perspective deformation in a simple application that computes the amount of available information for the tracking of a camera from a cloud of points.

    \subsection{Geometric covariance}  \label{subsec:geoCov}
        
Figure \ref{fig:geoVal} gives an overview of the simulated setup we use to assess the deformation model. We generate a set of random cameras, points and surfaces to produce a massive number of projection samples ($\approx10^6$). To simulate the perspective deformation suffered by planar patches, we add a small Gaussian noise to the reference coordinates of the points $\textbf{u}$, and measure the covariance of the projected error distribution. Then, for each projection, we obtain a covariance matrix $\bar{\textbf{C}}_{sim}$ analogous to the left Cauchy-Green deformation tensor in Equation \eqref{eq::leftCauchyTensor}. Figure \ref{subfig::montecarloVal} shows the relation between the simulated 2d deformation $\varepsilon^2_{sim} = \det (\bar{\textbf{C}}_{sim})$, and our estimation with the derivation in Section \ref{sec:perspDef}. 

\newcommand\figureScale{0.17}
\newcommand\figureWidth{0.14}

\begin{figure*}[t]
\centering
 \begin{subfigure}[b]{\figureWidth\textwidth}
  \includegraphics[scale=\figureScale]{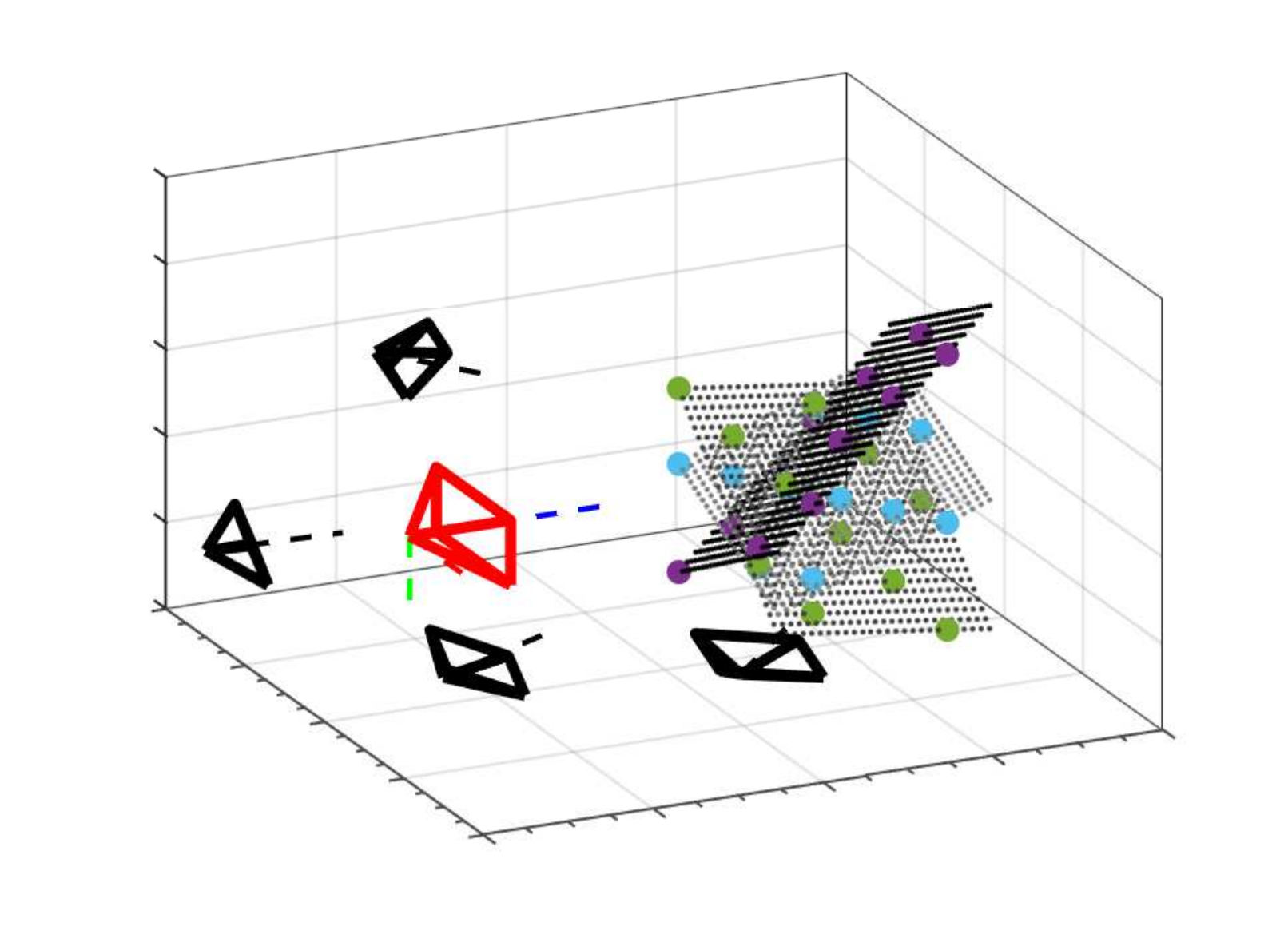}
  \caption{Planar}
  \label{fig:planar}
\end{subfigure}
\begin{subfigure}[b]{\figureWidth\textwidth}
  \includegraphics[scale=\figureScale]{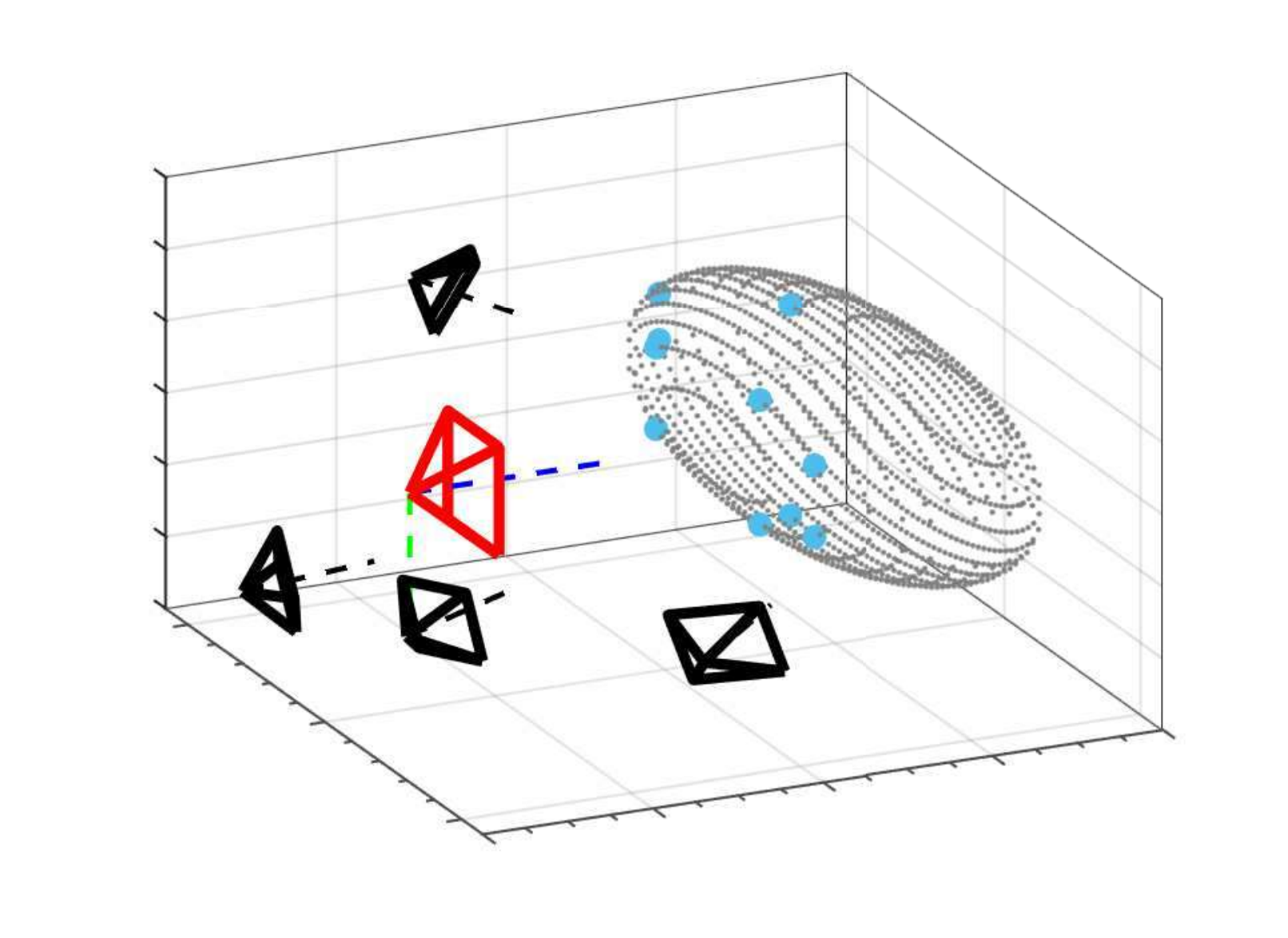}
  \caption{Ellipsoid}
  \label{fig:ellipsoid}
\end{subfigure}
\begin{subfigure}[b]{\figureWidth\textwidth}
  \includegraphics[scale=\figureScale]{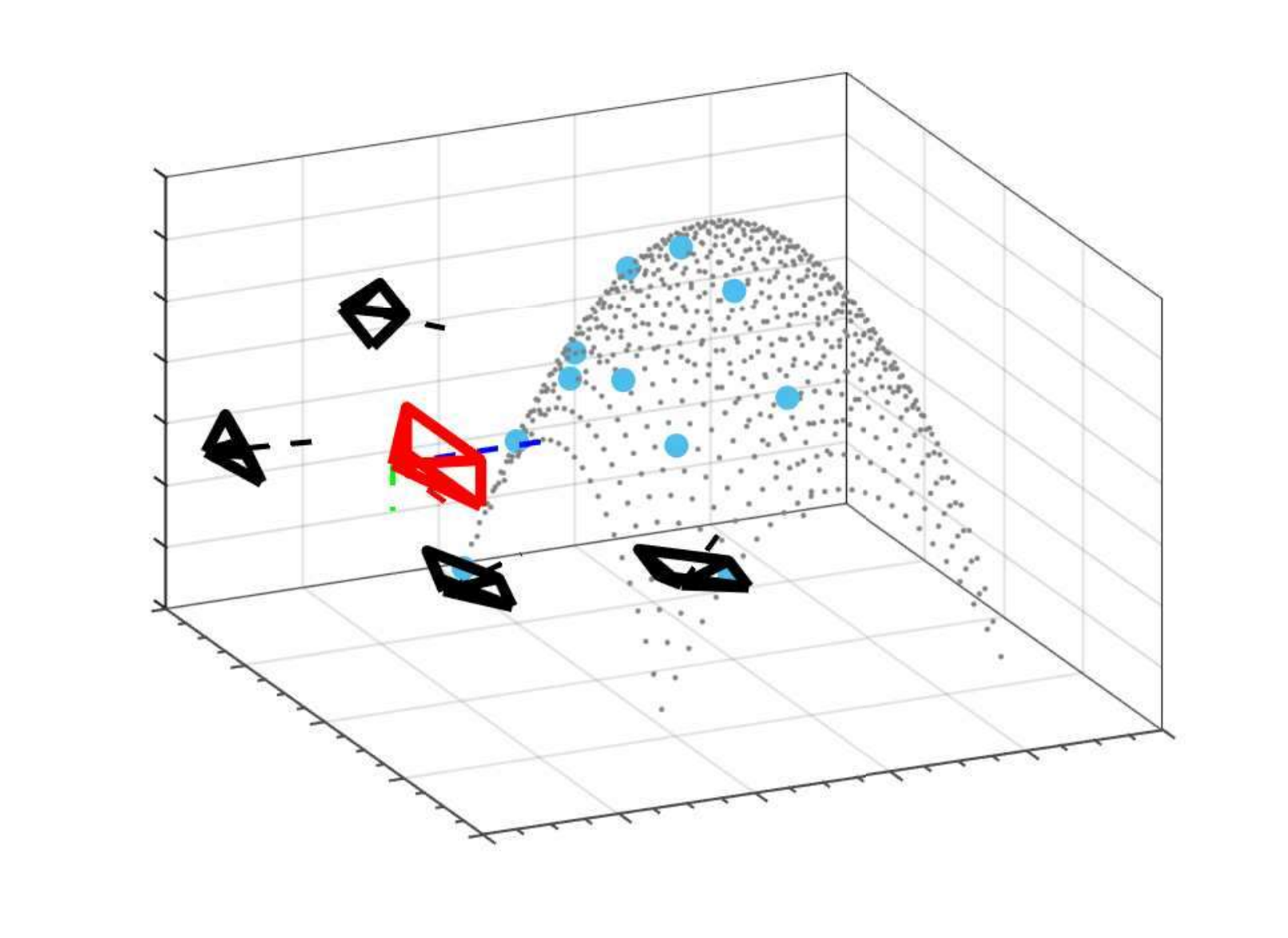}
  \caption{Elliptic par.}
  \label{fig:illiptic}
\end{subfigure}
\begin{subfigure}[b]{\figureWidth\textwidth}
  \includegraphics[scale=\figureScale]{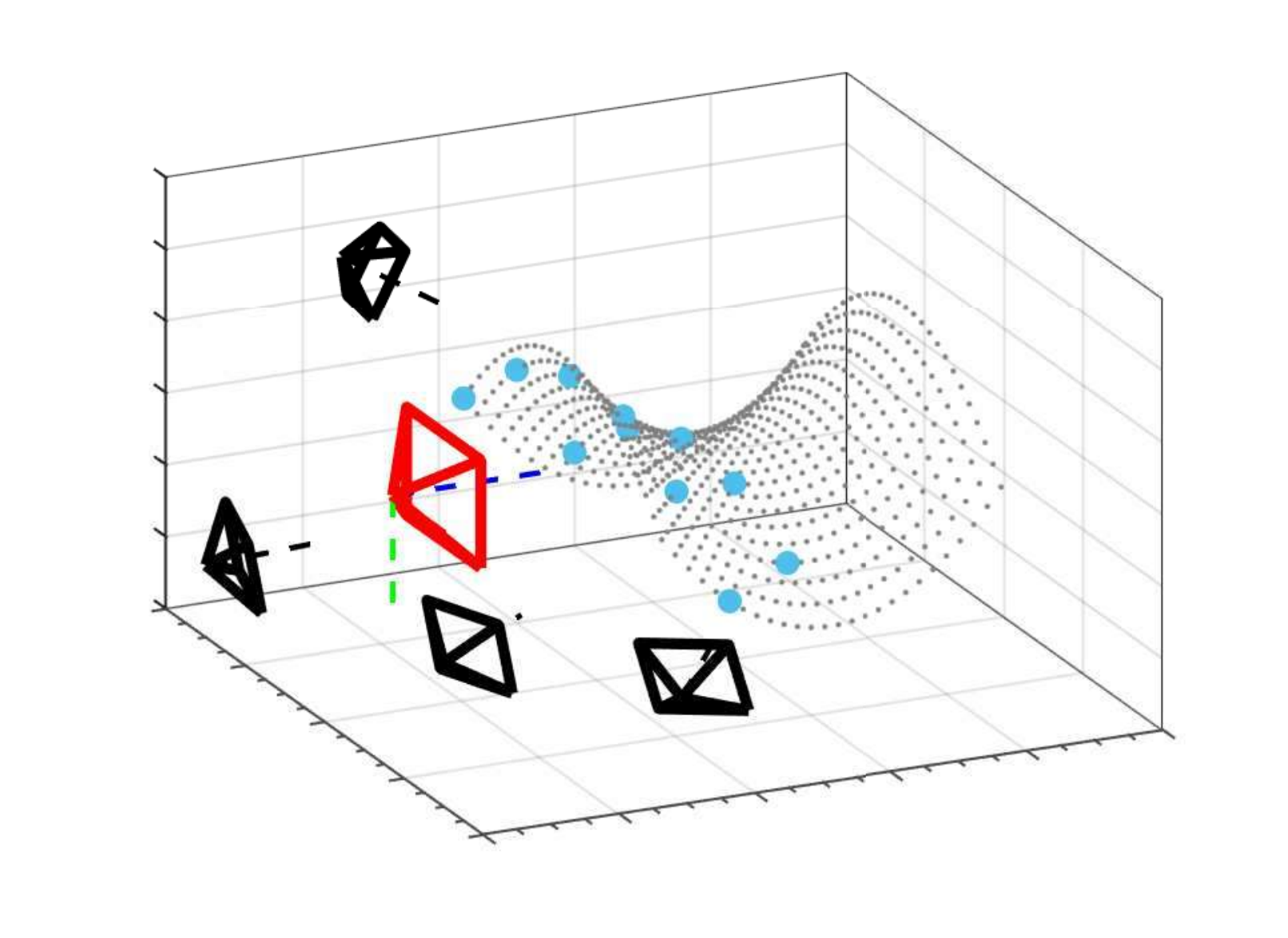}
  \caption{Hyperbolic par.}
  \label{fig:hyperbolic}
\end{subfigure}
\begin{subfigure}[b]{\figureWidth\textwidth}
  \includegraphics[scale=\figureScale]{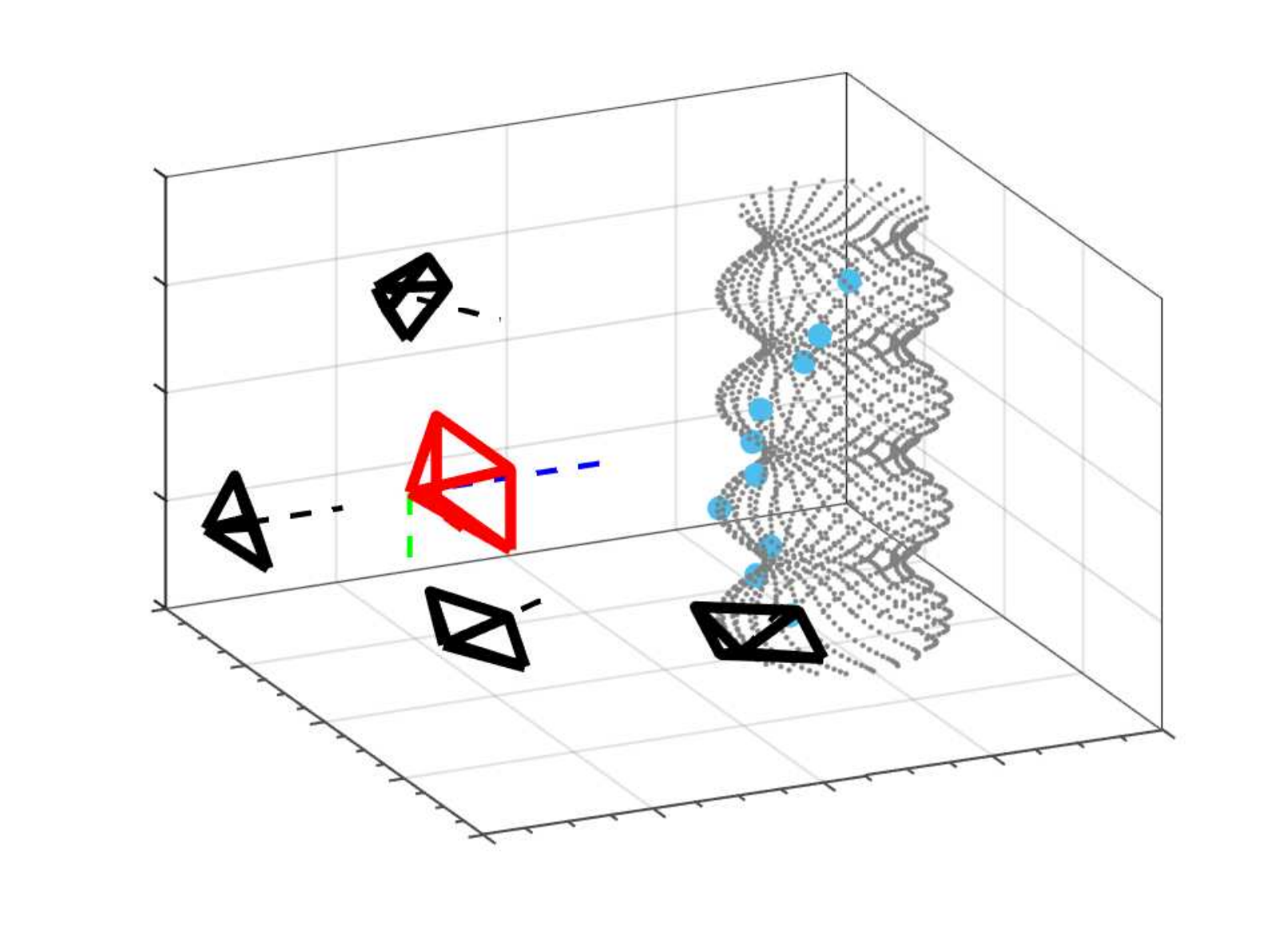}
  \caption{Rev. Sin.}
  \label{fig:revolution}
\end{subfigure}
\begin{subfigure}[b]{\figureWidth\textwidth}
  \includegraphics[scale=\figureScale]{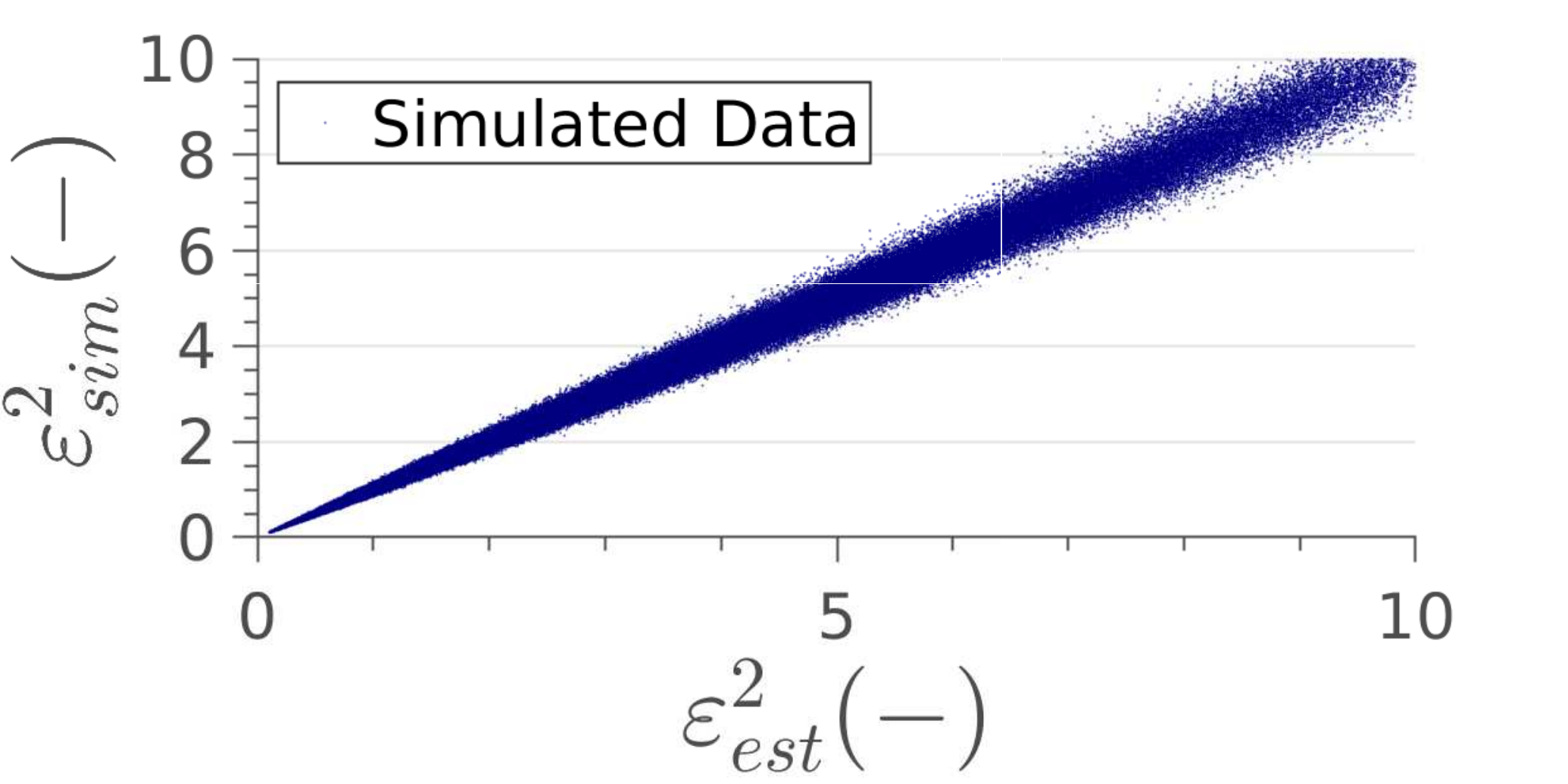}
  \caption{$\varepsilon^2_{est}$ vs. $\varepsilon^2_{sim}$.}
     \label{subfig::montecarloVal}
\end{subfigure}
\caption{{Monte Carlo validation}.  We show the comparison of our deformation estimation $\varepsilon^2_{est}$ with a simulation $\varepsilon^2_{sim}$ on a set of representative surfaces: planar, ellipsoid, elliptic and hyperbolic paraboloid, and a revolution sine. Figure \ref{subfig::montecarloVal} demonstrates how our model can be used, not only with planes, but with any parameterizable surface $z = S(\textbf{u})$ (see Section \ref{subsec:surface}). $^\dagger$ Cameras and points shown in this figure are just a subset chosen with visualization purposes. The total amount of point projections is $\approx10^6$.}
\label{fig:geoVal}
\end{figure*}

    \subsection{Photometric patches}  \label{subsec:photoPatches}

To ensure that only the influence of perspective deformation is being considered ($\sigma^2_\varphi = \sigma^2_\varepsilon$), we validate our model for planar photometric patches in the synthetic ICL-NUIM dataset \cite{handa2014benchmark}, which provides RGB-D sequences and the ground truth values for the camera and scene parameters.  

As in \cite{molton2004locally}, we consider patches as locally planar regions on 3D world surface instead as 2D templates in image space or more complex surfaces, since our formulation allows patches to be placed on any parametric surface (see Section \ref{subsec:geoCov}). Although salient points often appear at discontinuities, it is commonly assumed that it is possible to find a locally dominant plane for their representation (see Figure \ref{fig:intro}).

\begin{figure}
\centering
\begin{subfigure}[b]{.45\linewidth}
\includegraphics[width=\linewidth]{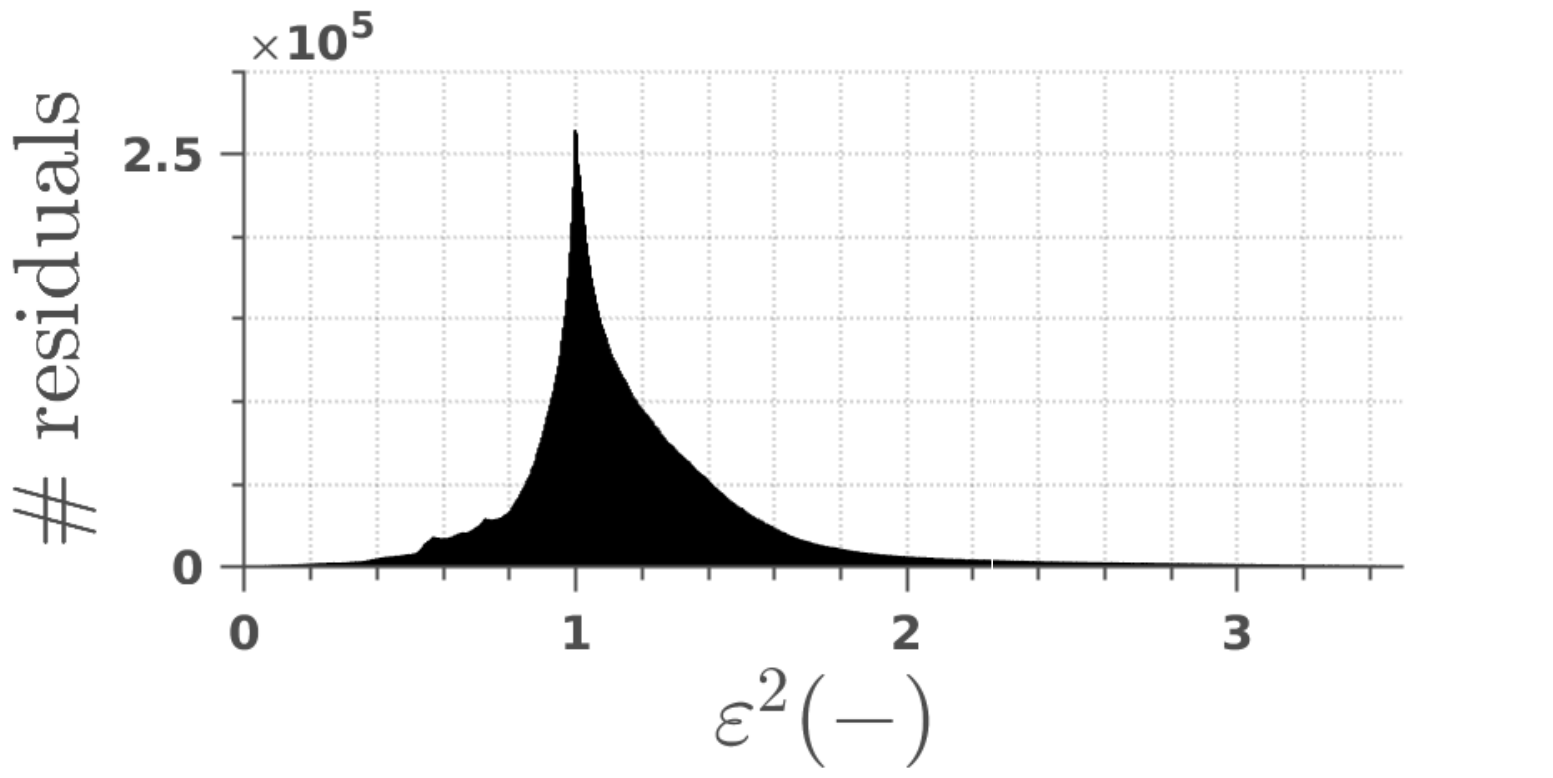}
\caption{Deformation data.}
\label{fig:defValData1}
\end{subfigure}
\begin{subfigure}[b]{.45\linewidth}
\includegraphics[width=\linewidth]{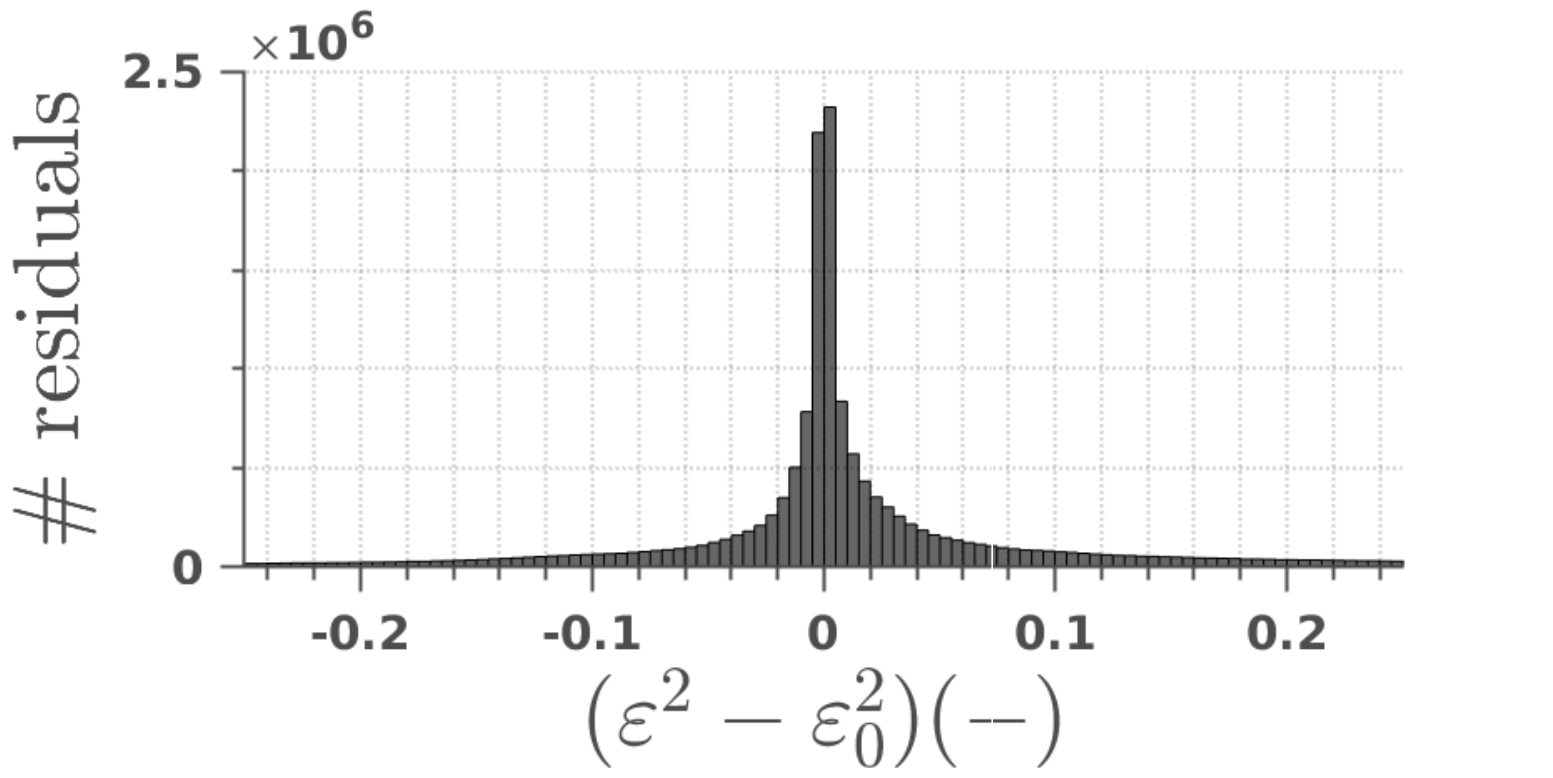}
\caption{Normalized photo. errors.}
\label{fig:defValData2}
\end{subfigure}

\begin{subfigure}[b]{.9\linewidth}
\includegraphics[width=0.9\linewidth]{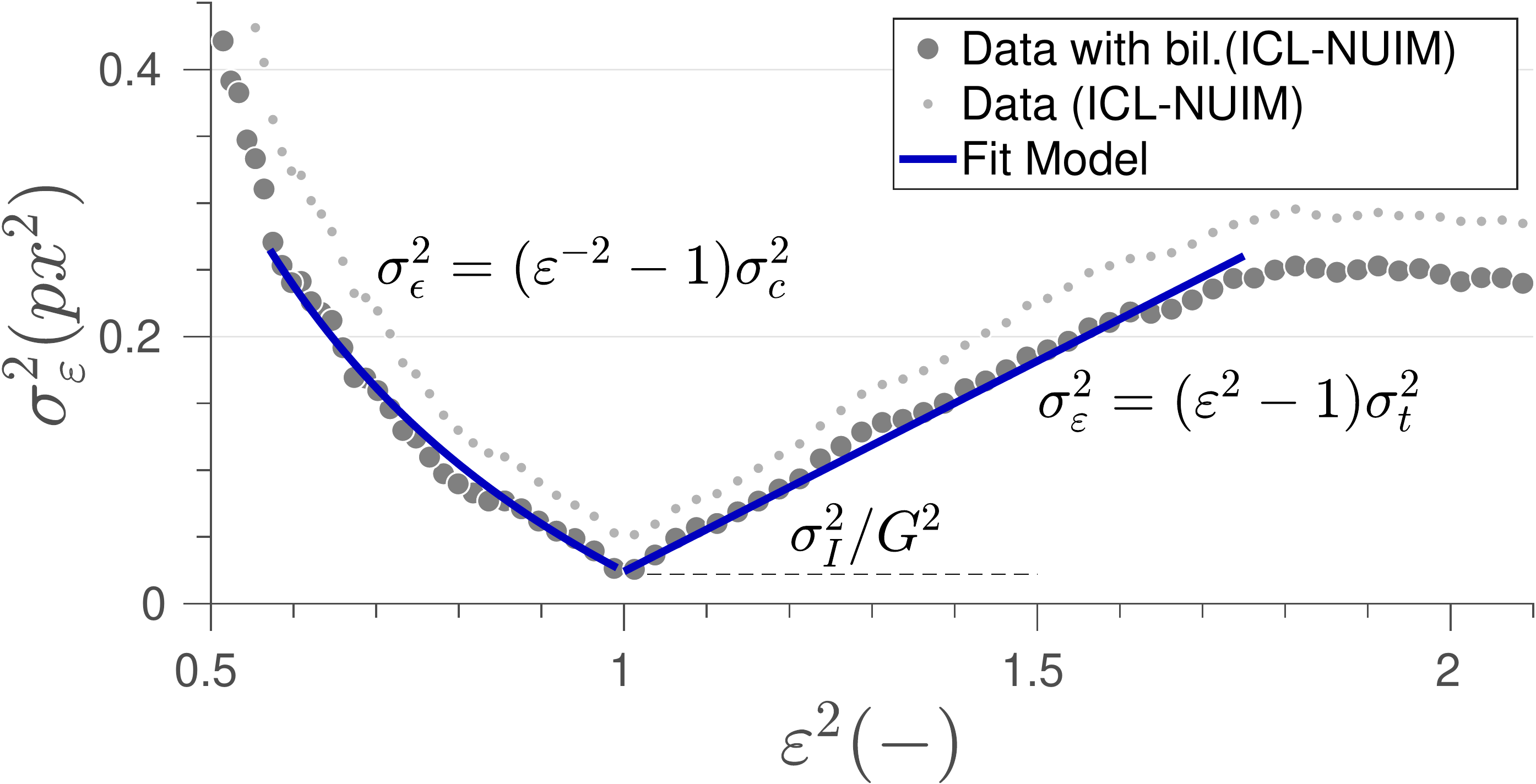}
\caption{\textbf{Photometric covariance model.} Results from the validation in the ICL-NUIM  dataset \cite{handa2014benchmark} show that the covariance for photometric patches increases along with the perspective deformation according to Equation \eqref{eq:geoCovSplit}.}
\label{fig:def2VsSigma2}
\end{subfigure}
\caption{\textbf{Photometric model validation}.}
\label{fig:animals}
\end{figure}

We gather a massive number of high gradient pixel projections between all image pairs within the same sequence, and extract their photometric errors, intensity gradients and predicted deformations. Figures \ref{fig:defValData1} and \ref{fig:defValData2} show histograms illustrating the number of data points, the deformation range and photometric errors in the dataset.

We group photometric errors according to their deformation and normalize them with the intensity gradient. Finally, we compute the covariance of the errors within each cluster obtaining the value for $\sigma^2_x$ in Equation \eqref{eq:photoCov} 

\begin{equation}
\sigma^2_x = \frac{\sigma_{r}^2}{G^2} = \frac{\sigma_{N}^2}{G^2} + \sigma_{\varepsilon}^2(\varepsilon^2) . 
\label{eq:photoCov1}
\end{equation}

Figure \ref{fig:def2VsSigma2} shows a representative sample of the results (using a pattern of $9$ pixels), clearly confirming a relation between perspective deformation and visual covariances and how our model captures it accurately. We highlight three significant outcomes: \textbf{1)} expressing compressions as its homologous tractions ($\varepsilon_t^2 = \varepsilon_c^{-2}$, see Section \ref{sec:perspDef}) unifies the behaviour of both covariance responses ($\sigma_t^2 \sim \sigma_c^2 $).  \textbf{2)} Following Ockham's razor, we define $\sigma_t^2(\varepsilon^2 - 1)$ and $\sigma_c^2(\varepsilon^{-2} - 1)$ simply with constant values. \textbf{3)} From the minimum covariance value in the absence of perspective deformation ($\sigma_\varepsilon^2 = 1$) we can derive with Equation \eqref{eq:patchNoise} the photometric noise in the images $\sigma^2_I$. Note how as expected the use of bilinear interpolation for intensities reduces the image noise to around $\frac{4}{9}$.

\textbf{Patch patterns}. Table \ref{table:patchApproach} evaluates our model for photometric residuals in different patch patterns, as proposed in \cite{engel2017direct}. The most relevant outcomes are: \textbf{1)} Bigger patches act as photometric filters, reducing the image noise $\sigma_I$. \textbf{2)} Bigger patches experience a faster degradation of the performance under traction, but perform better than smaller patches under compression. And vice versa, smaller patches deteriorate faster under compression and handle traction better. This effect is easily recognizable observing the changes of the response function parameters ($\sigma_t^2$, $\sigma_c^2$) or the range of deformation $\varepsilon^2$ in Table \ref{table:patchApproach} .     

The last columns of Table \ref{table:patchApproach} show results assuming the surface is perpendicular to the optical axis ($S_0$) or the backprojected ray ($S_\perp$). We observe that now traction effects are mostly dominated by the \textbf{surface assumption}, \emph{i.e.}, as the camera moves towards the points, depth inconsistencies arise. Yet, bigger patches still work better under compression.

\begin{table}
    \quad  
	\begin{minipage}{0.65\linewidth}
		\label{tab:le}
		\centering
		\resizebox{\textwidth}{!}{%
		\begin{tabular}{ c c c c c } 
             \multicolumn{1}{c}{}&\multicolumn{4}{c}{$S\{\mathtt{R}^2 > 0.9975\}$}\\
             \cline{1-5}
             radius& $\sigma_{I}$&$\sigma_{t}^2$&$\sigma_{c}^2$&$\varepsilon^2$\\
             \cline{1-5}
             $0.5$        &\color{myred}4.02&\color{mygreen}0.30&\color{myred}0.49&\color{myred}0.50\color{black}-\color{mygreen}1.75\\
             $1$          &2.69&0.31&0.43&0.48-1.70 \\
             $\sqrt{2}$   &2.66&0.32&0.40&0.46-1.66\\
             \color{blue}\textbf{2}&\color{blue}\textbf{2.25}&\color{blue}\textbf{0.33}&\color{blue}\textbf{0.35}&\color{blue}\textbf{0.43-1.61}\\
             $2\sqrt{2}$          &1.99&0.35&0.31&0.39-1.52\\
             $4$                  &1.74&0.36&0.28&0.33-1.41\\
             $4\sqrt{2}$          &\color{mygreen}1.64&\color{myred}0.38&\color{mygreen}0.22&\color{mygreen}0.25\color{black}-\color{myred}1.25 \\
            \end{tabular}
        }
	\end{minipage}
	\quad 
	\begin{minipage}{0.1\linewidth}
		\centering
		\vspace{0.5cm}
		\label{figu:re}
		\includegraphics[scale=0.11]{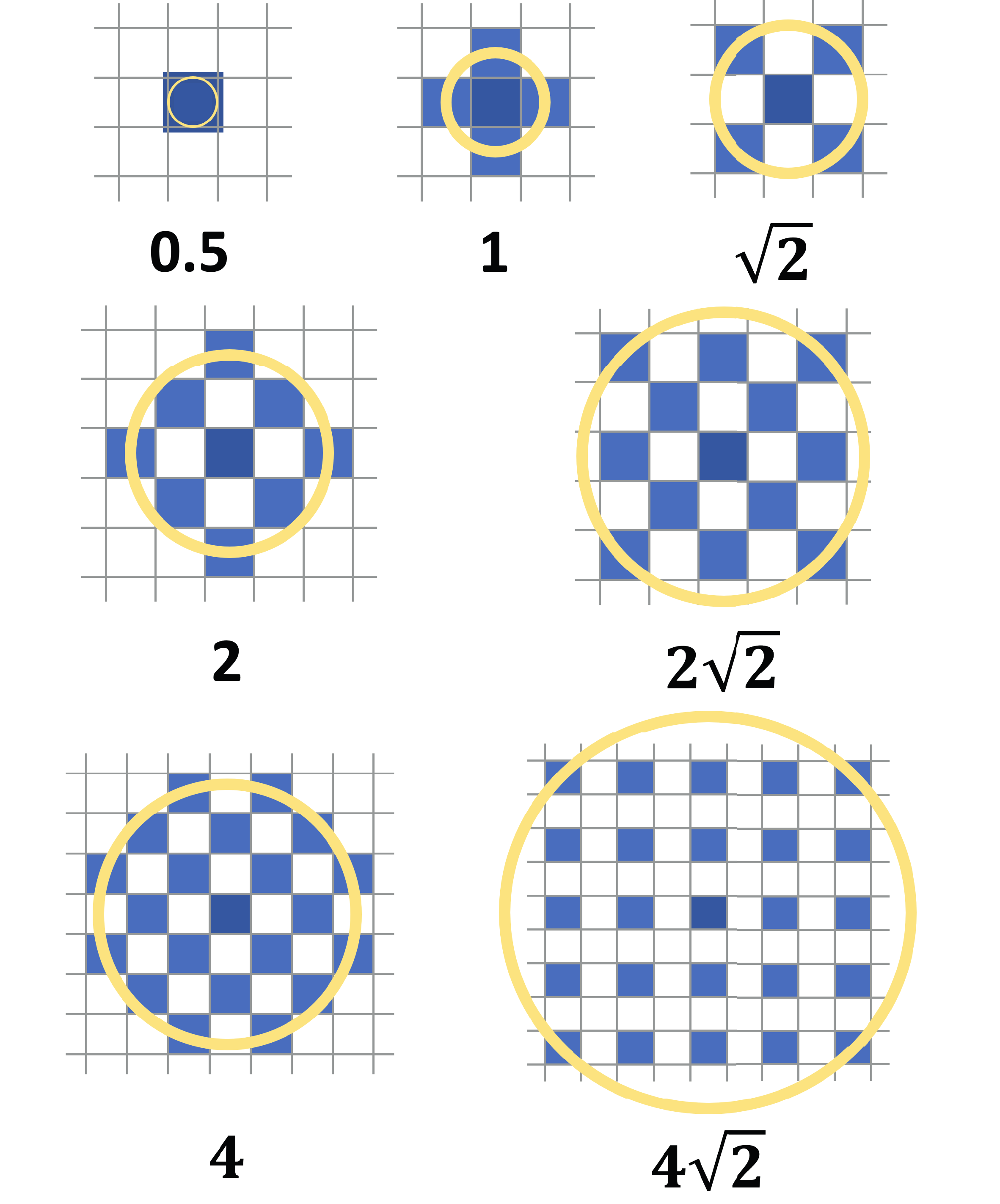}
	\end{minipage}
	\vspace{-0.2cm}
	\caption{\textbf{Photometric model fitting.} For each patch size, we obtain the parameters of Equation \eqref{eq:geoCovSplit} that maximize the coefficient of determination $\mathtt{R}^2$ within a certain range of deformation $\varepsilon^2$. Per-column best is displayed in green, worst in red and balanced in blue. Small patches behave better for traction, large patches for compression.}
	\label{table:patchApproach}
\end{table}

\newcommand\figureScale{0.105}
\newcommand\figureWidth{0.17}

\begin{figure*}
\begin{subfigure}[t]{0.12\textwidth}
\includegraphics[scale=0.08]{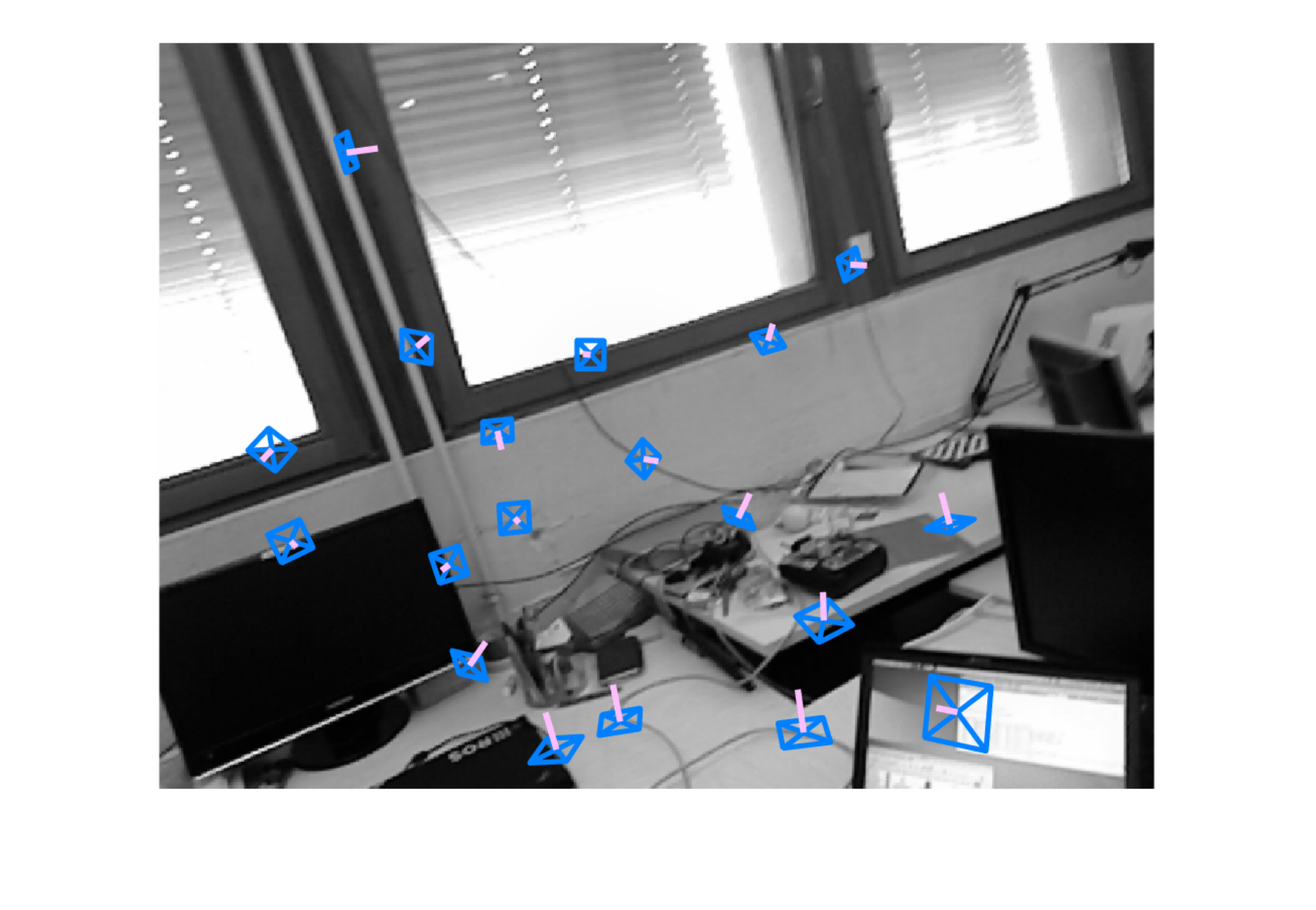}
\caption{Fr1.}
\end{subfigure}
\begin{subfigure}[t]{\figureWidth\textwidth}
\includegraphics[scale=\figureScale]{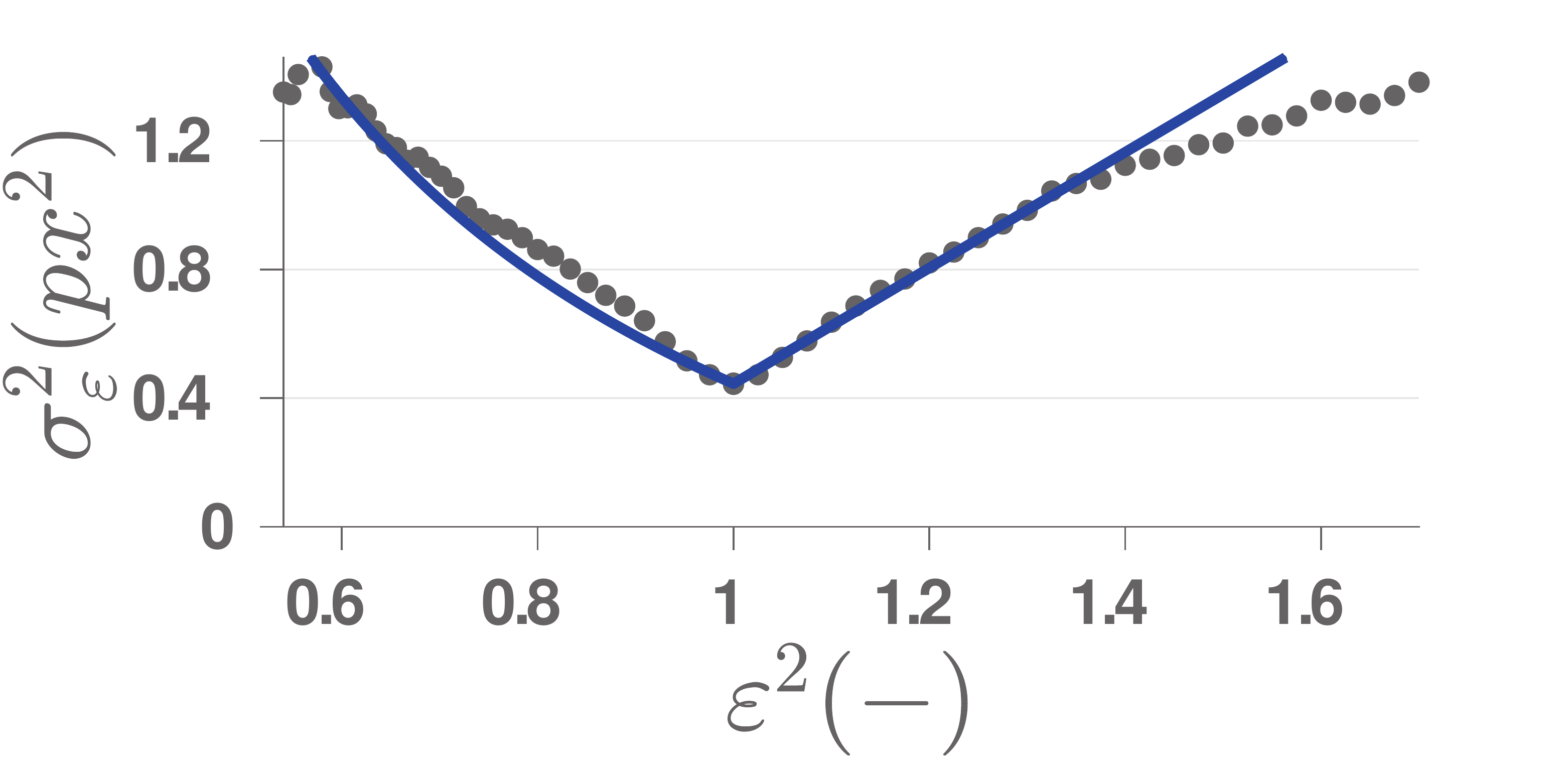}
\caption{Photo. Model}
\end{subfigure}
\begin{subfigure}[t]{\figureWidth\textwidth}
\includegraphics[scale=\figureScale]{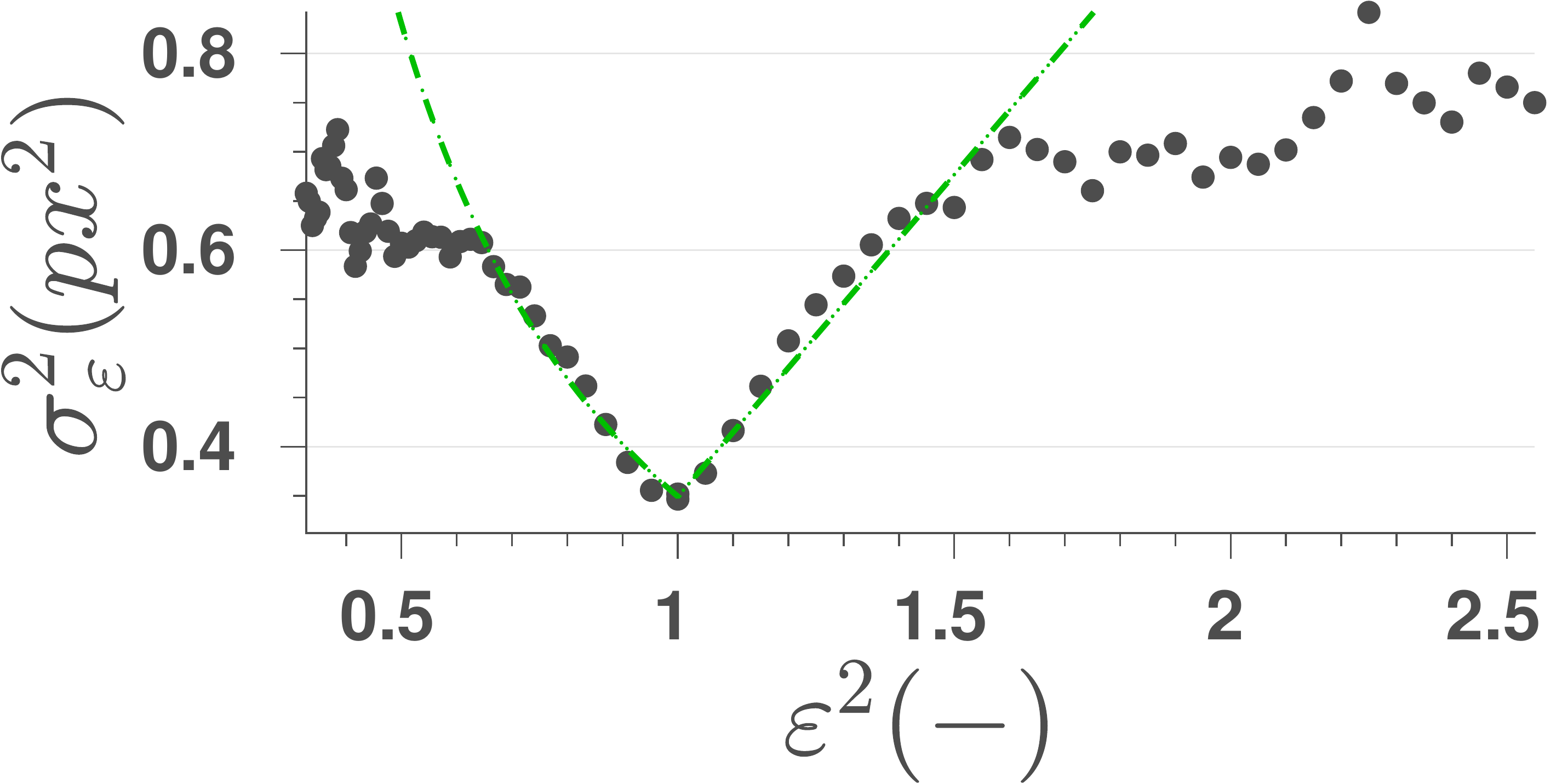}
\caption{ORB. Model}
\end{subfigure}
\begin{subfigure}[t]{0.12\textwidth}
\includegraphics[scale=0.08]{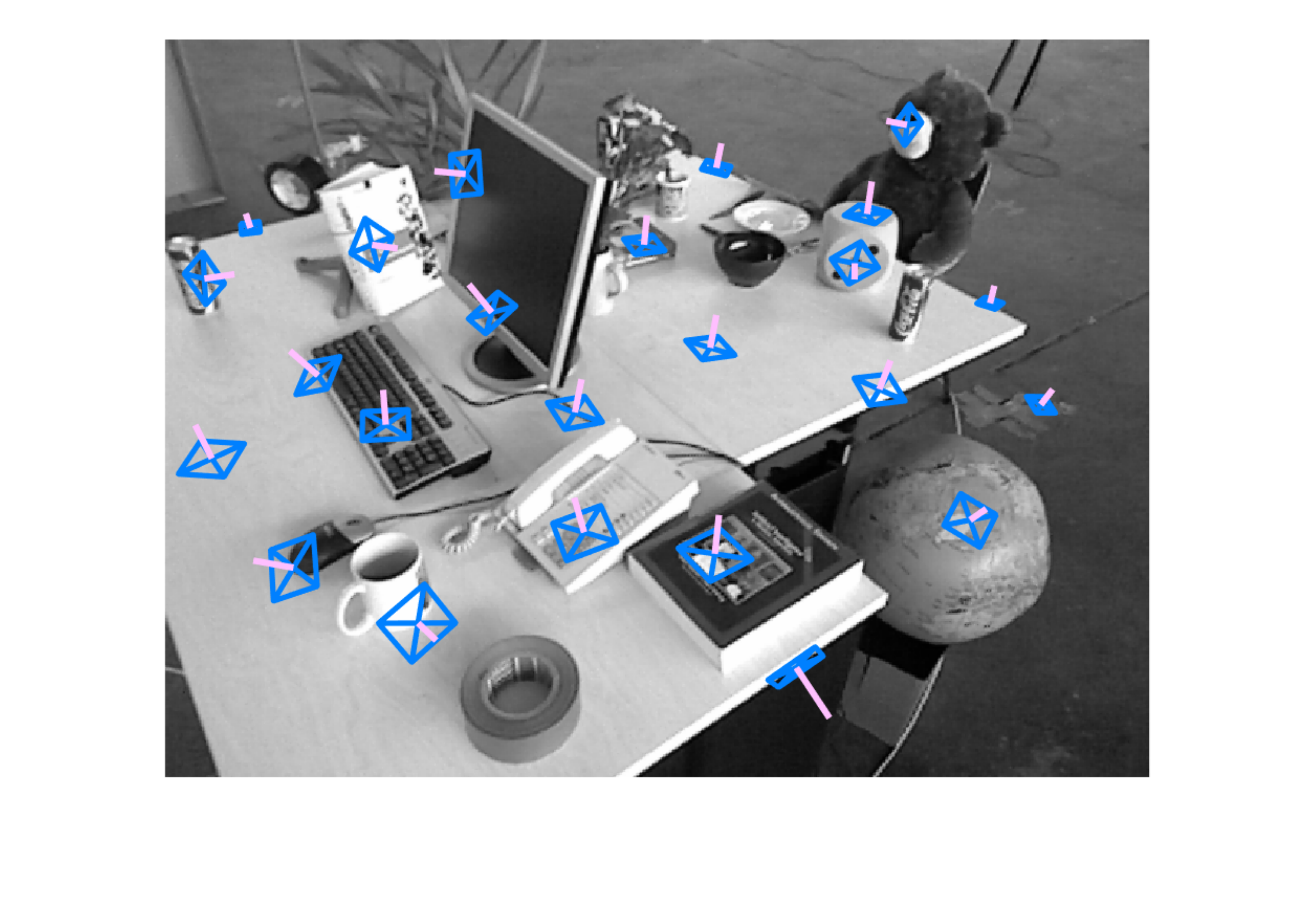}
\caption{Fr2.}
\end{subfigure}
\begin{subfigure}[t]{\figureWidth\textwidth}
\includegraphics[scale=\figureScale]{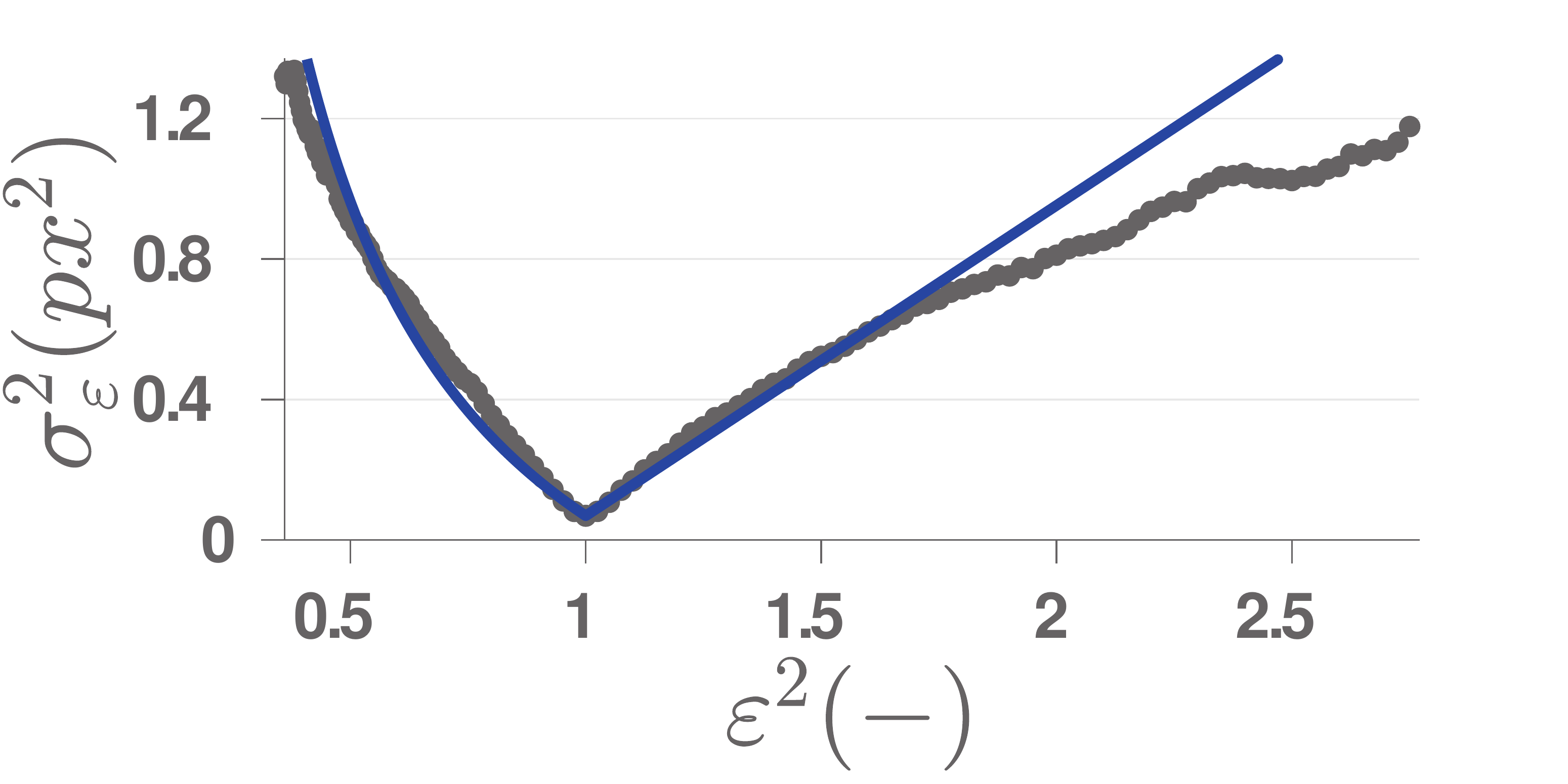}
\caption{Photo. Model}
\end{subfigure}
\begin{subfigure}[t]{\figureWidth\textwidth}
\includegraphics[scale=\figureScale]{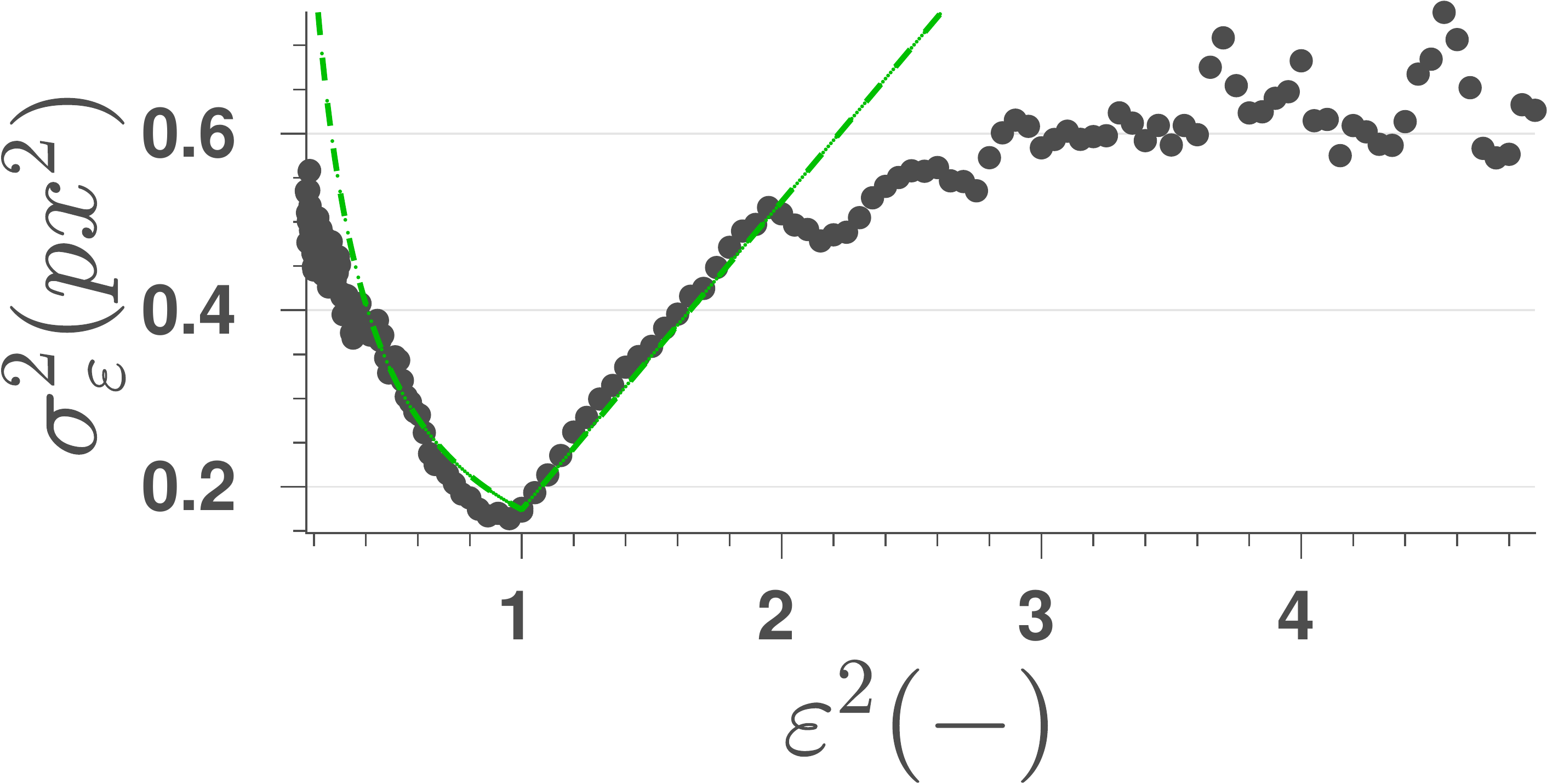}
\caption{ORB. Model}
\end{subfigure}

\begin{subfigure}[t]{0.12\textwidth}
\includegraphics[scale=0.08]{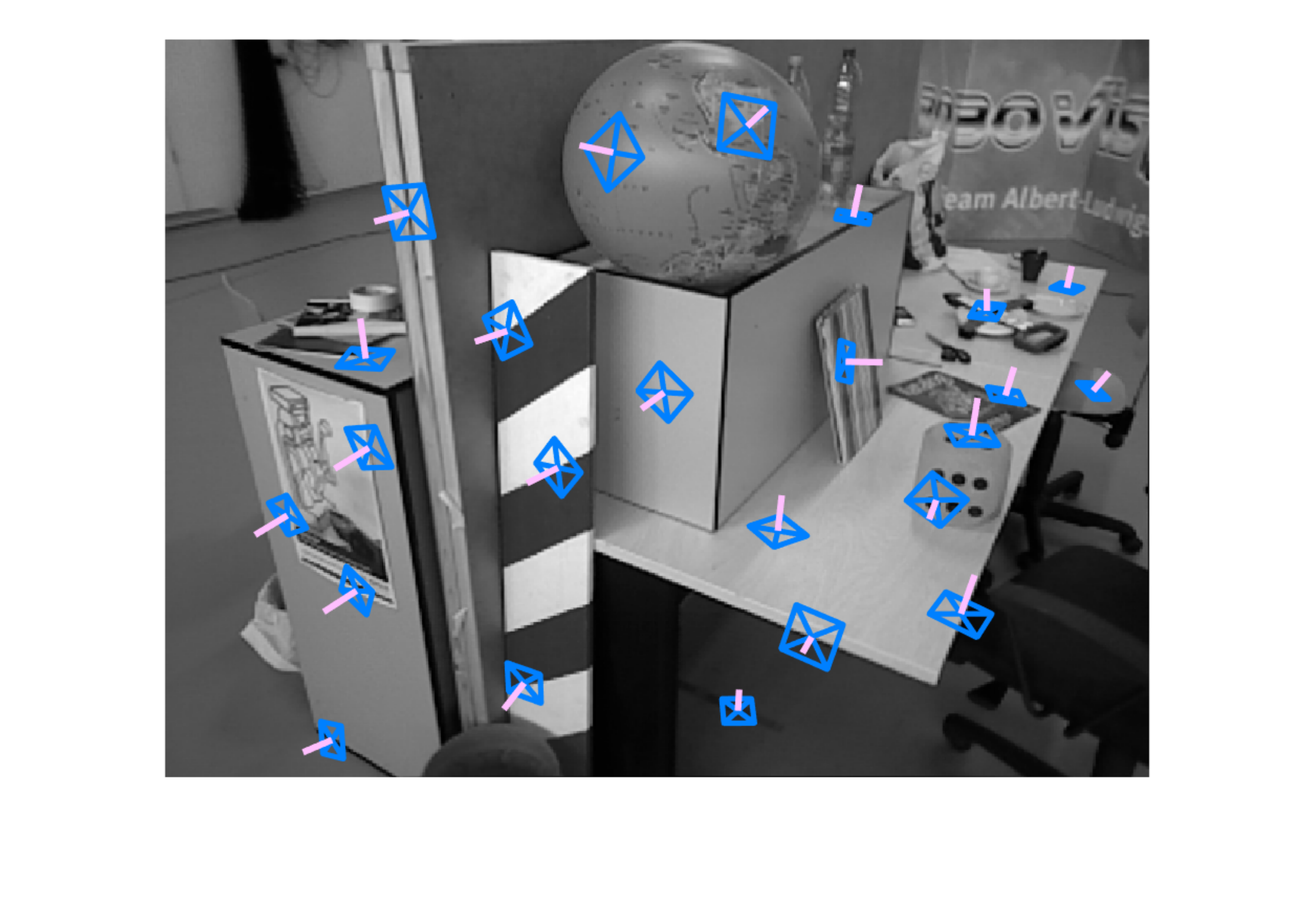}
\caption{Fr3.}
\end{subfigure}
\begin{subfigure}[t]{\figureWidth\textwidth}
\includegraphics[scale=\figureScale]{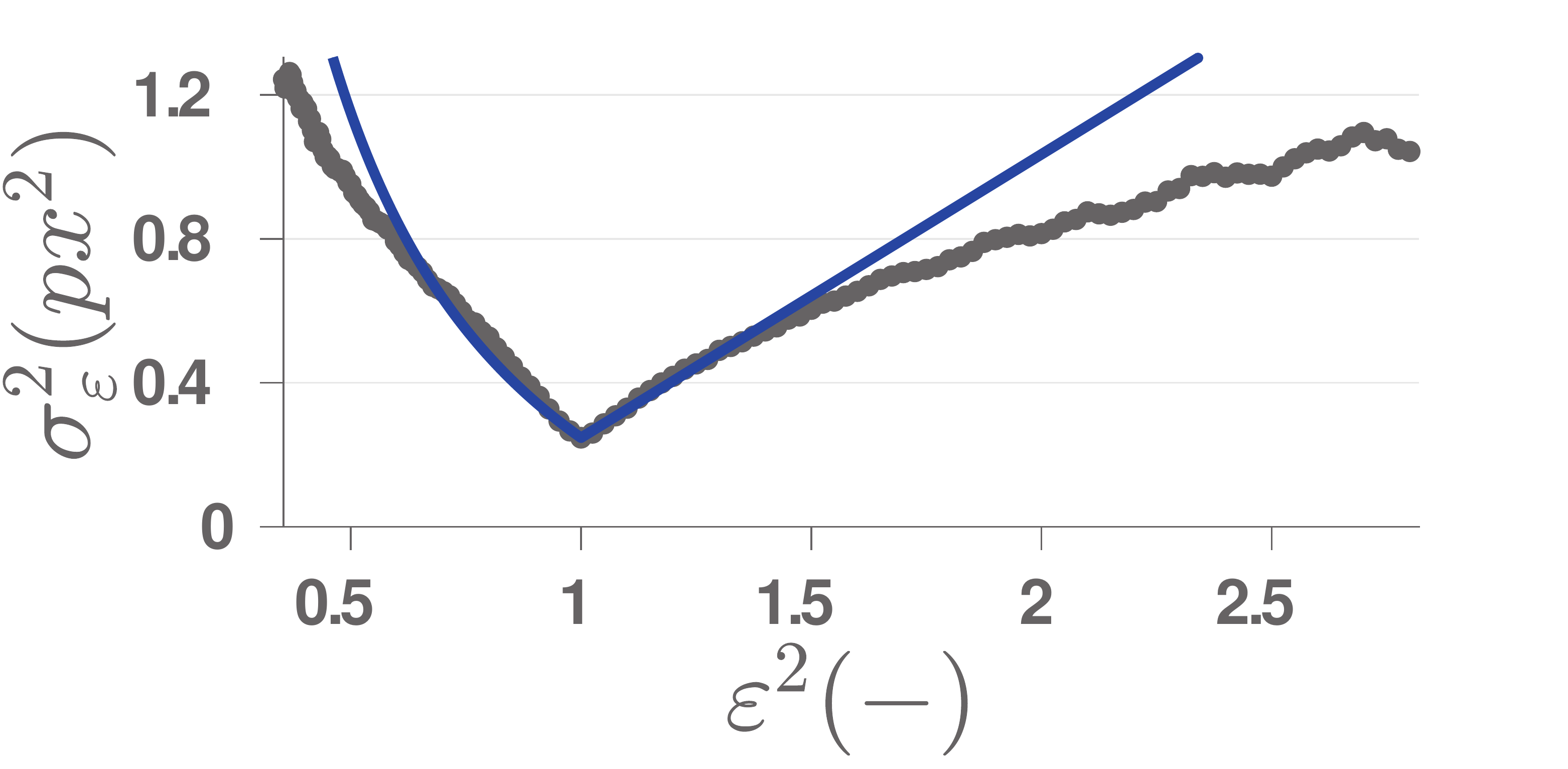}
\caption{Photo. Model}
\end{subfigure}
\begin{subfigure}[t]{\figureWidth\textwidth}
\includegraphics[scale=\figureScale]{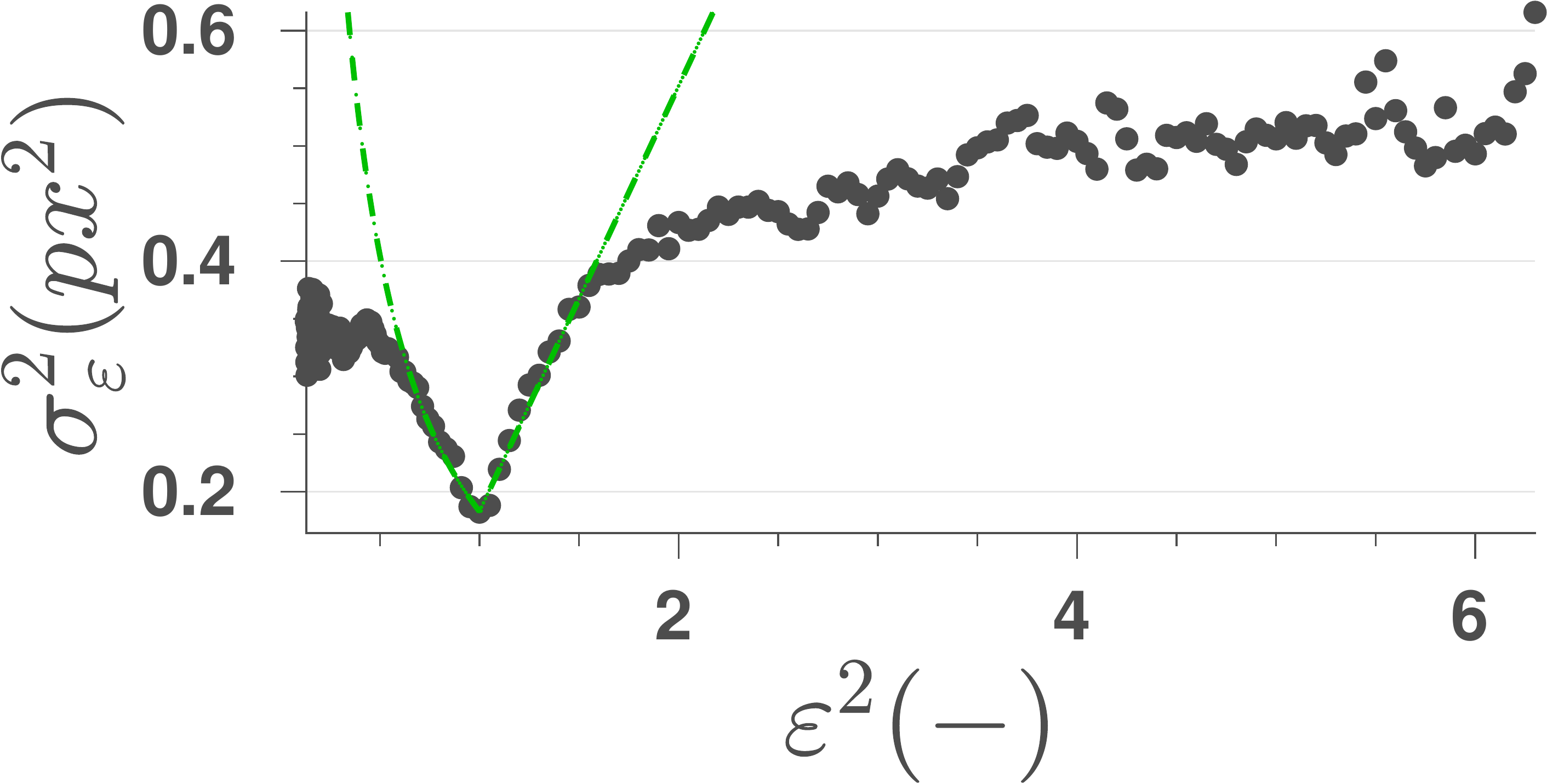}
\caption{ORB. Model}
\end{subfigure}
\begin{subfigure}[t]{0.12\textwidth}
\includegraphics[height=1.4cm,width=2.3cm]{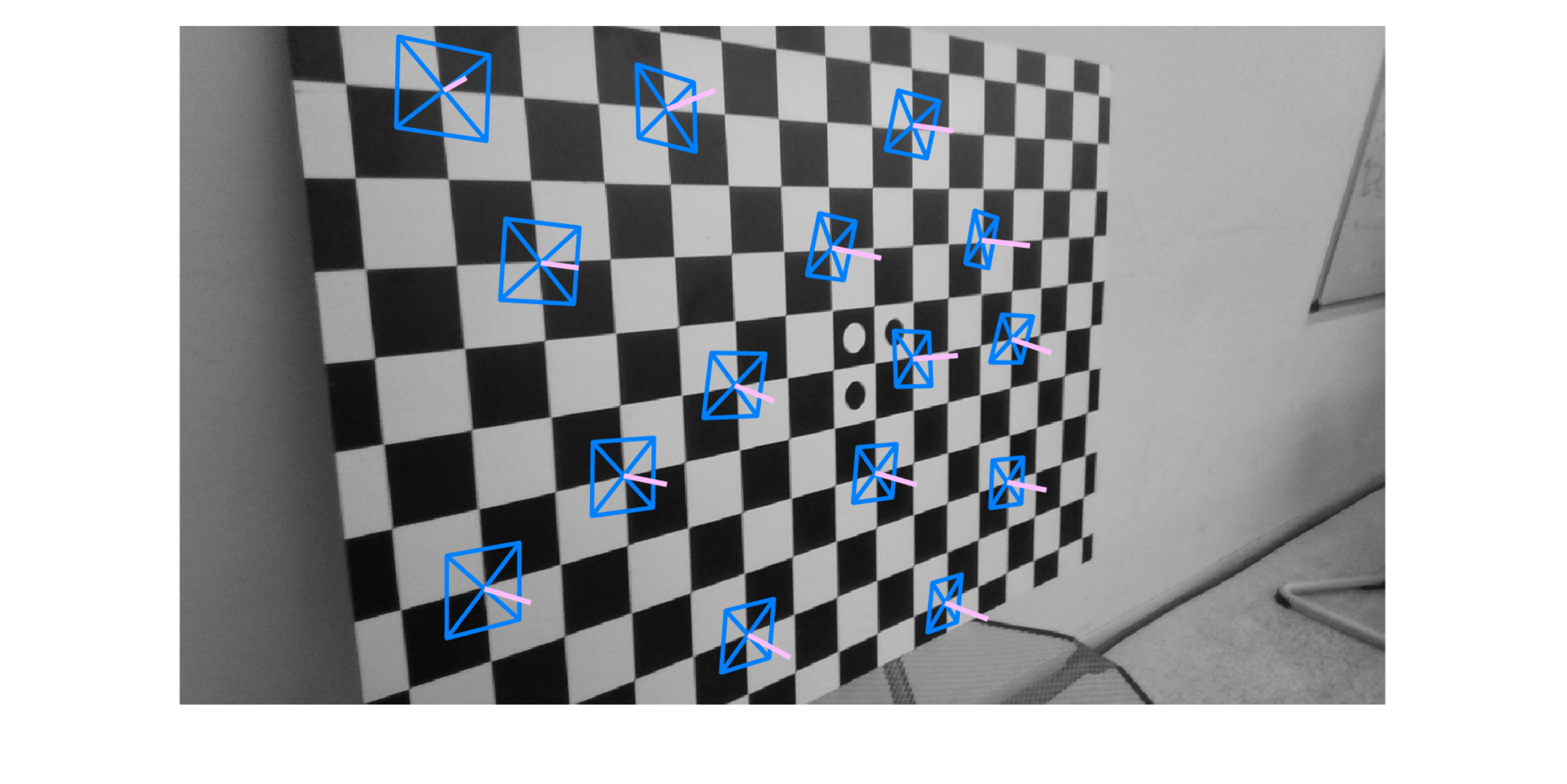}
\caption{realsense}
\end{subfigure}
\begin{subfigure}[t]{\figureWidth\textwidth}
\includegraphics[scale=\figureScale]{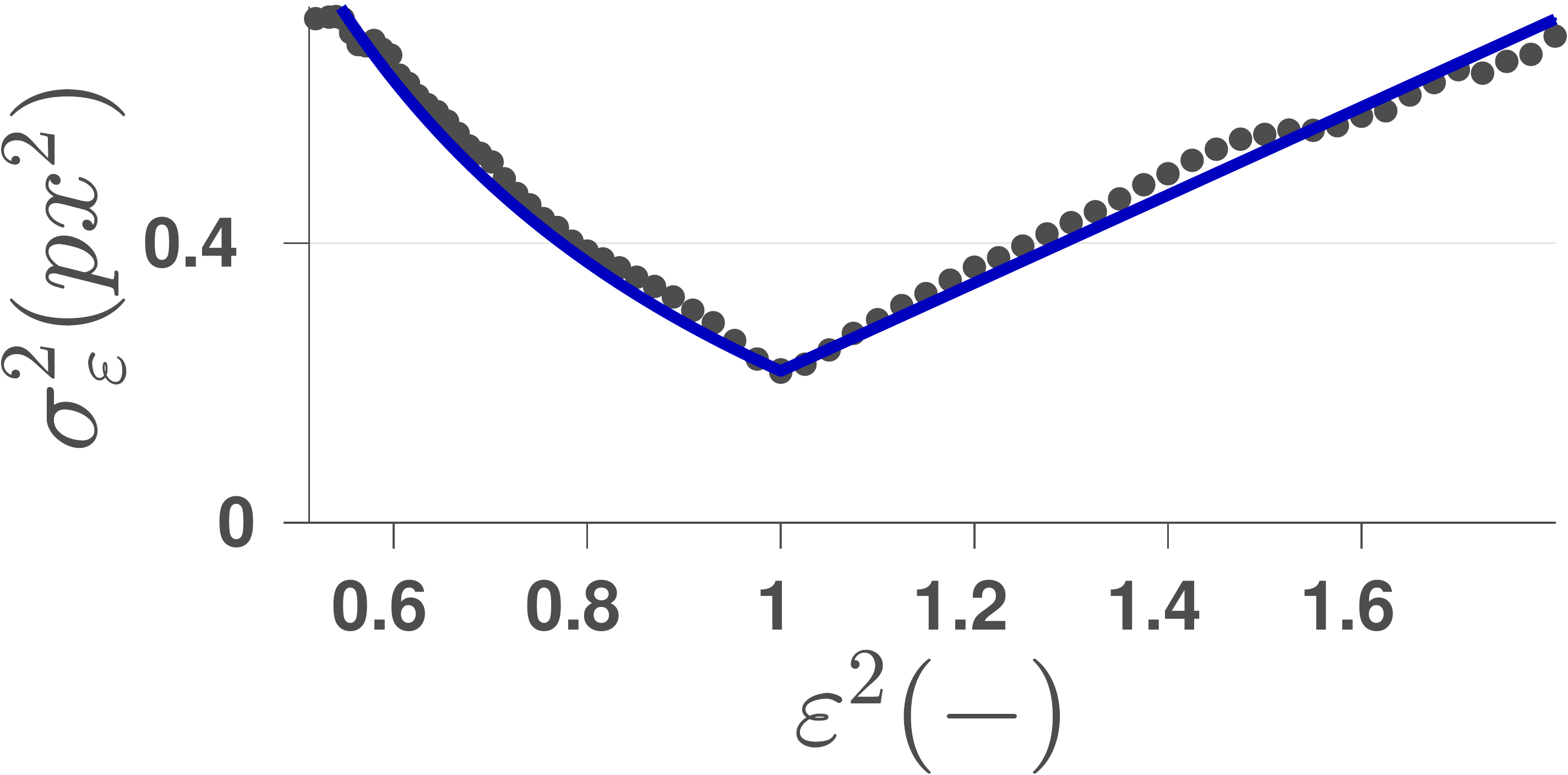}
\caption{Photo. Model}
\label{subfig:realsensePhoto}
\end{subfigure}
\begin{subfigure}[t]{\figureWidth\textwidth}
\includegraphics[scale=\figureScale]{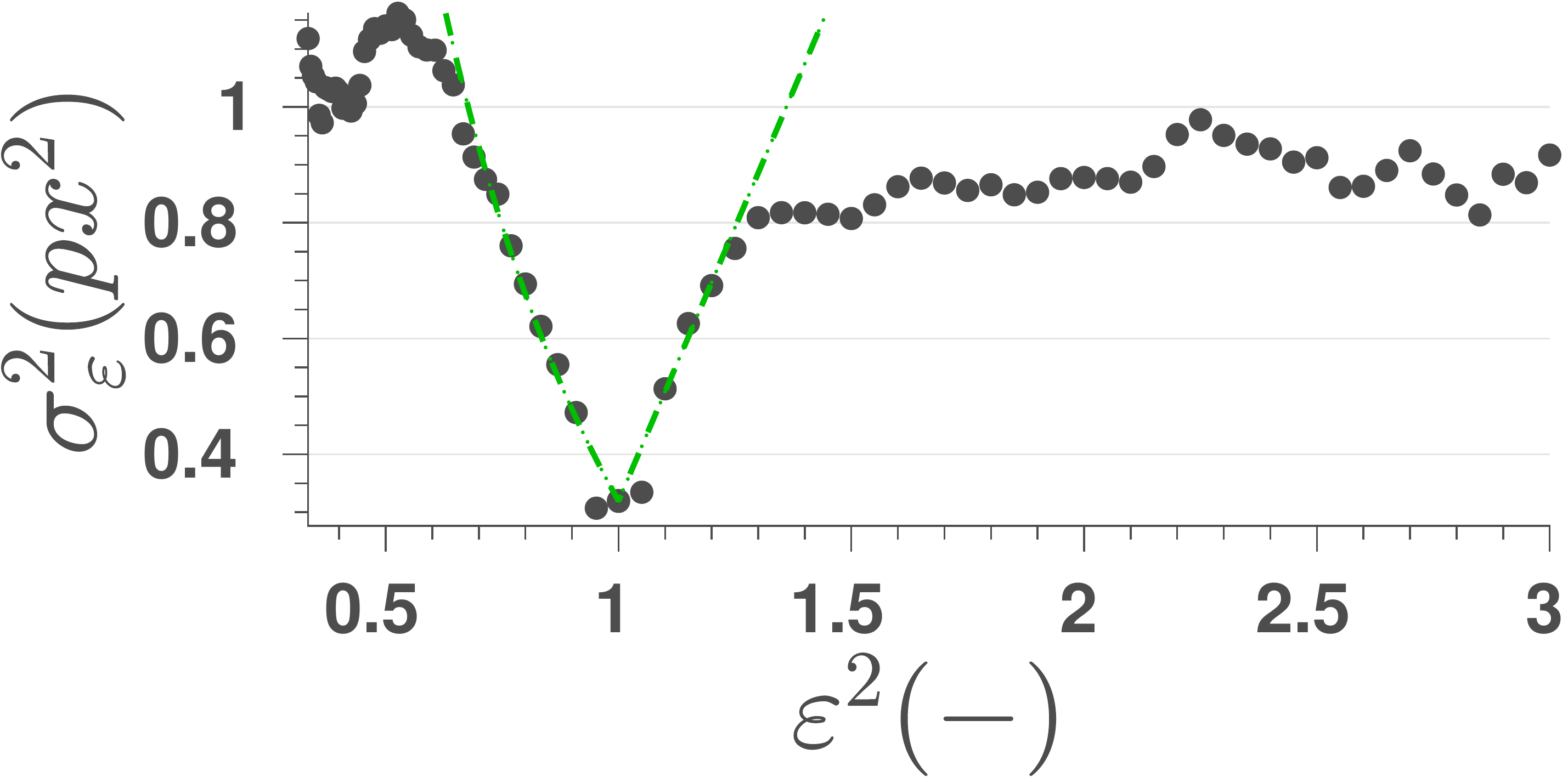}
\caption{ORB. Model}
\end{subfigure}

\caption{\textbf{Model validation with real data}. We perform the model fitting of Equation \eqref{eq:geoCovSplit}, described in sections \ref{subsec:photoPatches} and \ref{subsec:featMethods}, with data from three cameras of the sequences of the RGB-D TUM dataset \cite{sturm12iros} and with data recorded with a realsense D435i depth camera. It can be seen how our covariance model for photometric patches and ORB features fits the noise distribution of the four cameras.}
\label{fig:photoCal_fr}
\end{figure*}

\begin{table}[h]
\centering
\begin{tabular}{c c c c c c c} 
camera & $\sigma_I$ & $\sigma^2_t$ & \multicolumn{1}{c}{$\sigma^2_c$}& $\varepsilon^2 \in$& $\mathtt{R}^2$ & $\frac{\sigma^2_t}{\sigma^2_c}$\\ [0.5ex]
\hline
fr1       & 47.2 & 1.80 & 1.34 & 0.6-1.4 & 0.95 & 1.3 \\ 
fr2       & 22.0 & 0.88 & 0.89 & 0.6-1.8 & 0.97 & 1.0\\
fr3       & 34.4 & 0.79 & 0.90 & 0.7-1.4 & 0.98 & 0.9\\
realSense & 30.8 & 0.63 & 0.62 & 0.6-1.9 & 0.97 & 1.0\\
\end{tabular}
\caption{\textbf{Model validation for photometric 9-pixel patches with real data}. The table shows the parameters of Equation \eqref{eq:geoCovSplit} estimated for the cameras of the RGBD-TUM dataset (fr) and a realsense D435i depth camera. Note how the  9-pixel patch behaves similarly under traction and compression deformation.}
\label{table:photoCov_realData}
\end{table}

\textbf{Real Data}. We repeat the experiment in Figure \ref{fig:def2VsSigma2} with real data from three different cameras of the public RGB-D TUM dataset \cite{sturm12iros}. Figure \ref{fig:photoCal_fr} shows how again, our model accurately captures the visual covariance produced by perspective deformation of photometric planar patches. An important outcome from the real data with different cameras and sequences is the validation of the assumption of considering patches as locally planar regions rather than more complex surfaces. However, it also leads to saturation of the model under strong changes in perspective deformation. Finally, to have a cleaner experiment with data from a depth sensor but trying to minimize any source of noise, we validated the model with data from an intel realsense D435i depth camera facing a checkerboard pattern. Figure \ref{subfig:realsensePhoto} shows how our model fits more reliably in this validation performed under more suitable conditions. Table \ref{table:photoCov_realData} collects the coefficients of the model validation that we will use in the experiments of Section \ref{sec:experiments}.

    \subsection{Feature-based methods}  \label{subsec:featMethods}

\begin{figure}[t]
\centering
\begin{subfigure}[b]{0.23\textwidth}
    \includegraphics[width=\textwidth]{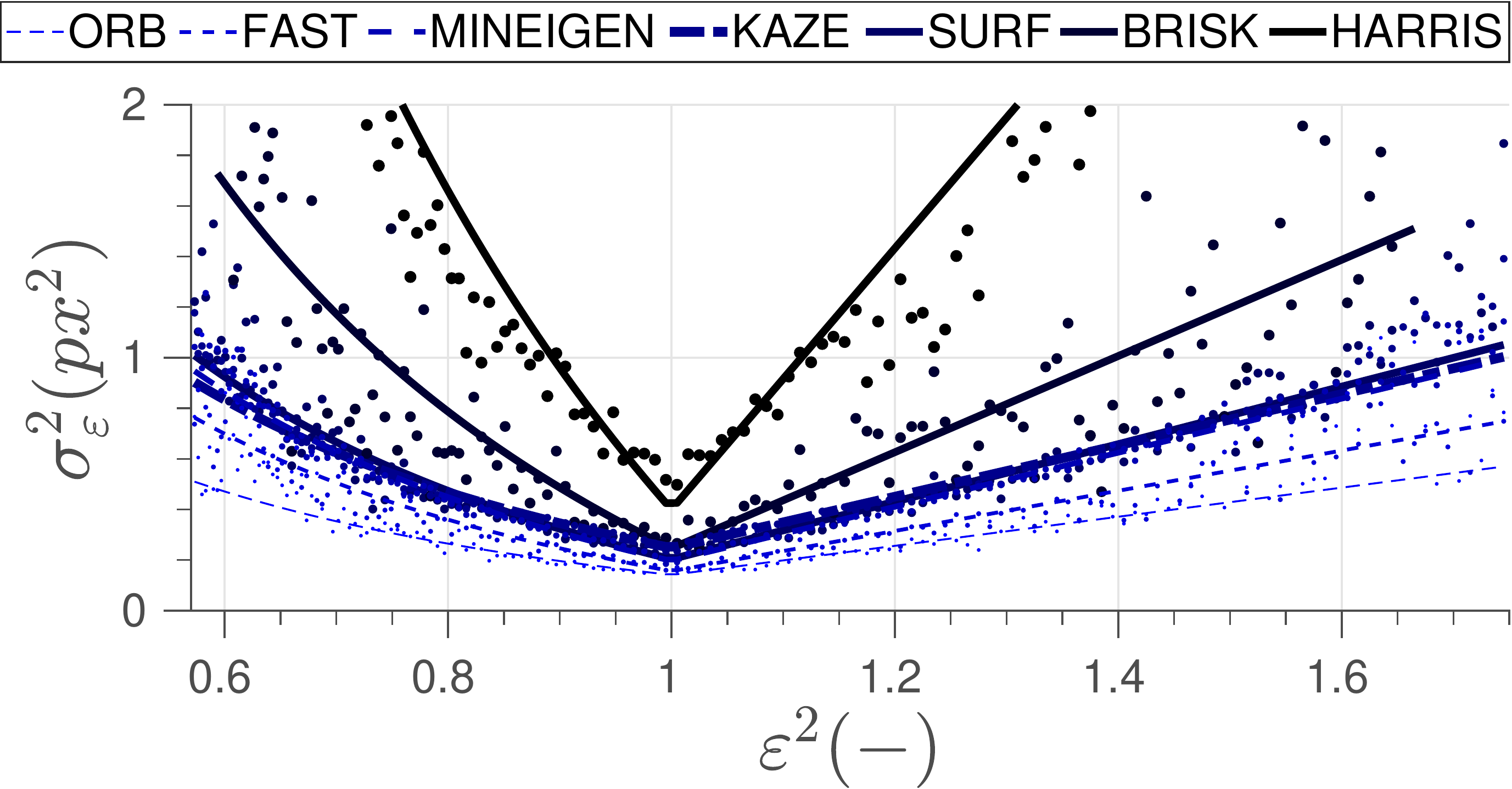} 
    \caption{Visual cov. for features.}
    \label{fig:features}
\end{subfigure}
\begin{subfigure}[b]{0.23\textwidth}
    \includegraphics[width=\textwidth]{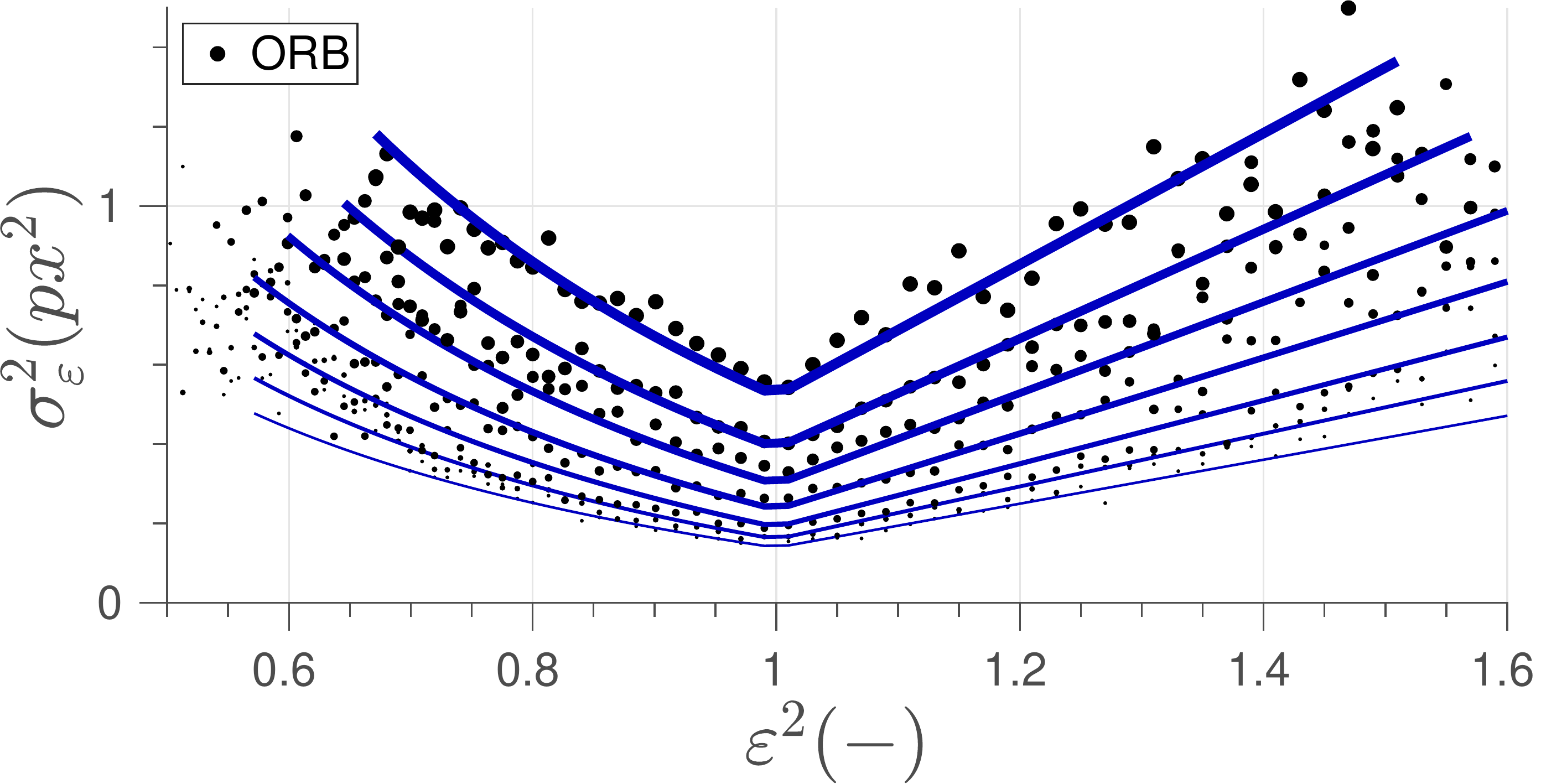}  
    \caption{Visual cov. for ORB.}
    \label{fig:orb}
\end{subfigure}
\caption{ In \ref{fig:features} all descriptors increase their covariance with the deformation. \ref{fig:orb} shows the deformation covariance of ORB depending of the scale of the feature. From bottom to top the resolution becomes coarser.}
\end{figure}

We keep the experimental setup of the previous section, but we now evaluate the geometric reprojection residual in feature-based methods. Similarly to Figure \ref{fig:def2VsSigma2}, we report in Figure \ref{fig:features} the dependency of the reprojection error with perspective deformation for different point features. The results show again a clear relation between the perspective deformation and the visual residual, and how our model fits reasonably the simulation data. 

Table \ref{table:geoCov_realData} shows the results for our model fitting. In the column titled $\sigma_t^2/\sigma_c^2$, it is relevant to note that visual covariances tend to grow faster for traction than compression. This is consistent with our photometric validation (see Table \ref{table:patchApproach}), where covariances in large patches grew faster under traction. Moreover, these results agree with \cite{yang2018challenges}, that shows experimentally the effect of motion bias in ORB-SLAM2 \cite{mur2017orb} and DSO \cite{engel2017direct}. They show a noticeable degradation for ORB-SLAM2 when the camera is moving forward, meaning that points mainly approach and consequently patches suffer from traction. On the other hand, DSO using photometric patches of radius $2$ (in our Table \ref{table:patchApproach}, with balanced traction and compression coefficients) does not show such bias. Our findings here are a step forward towards a more complete understanding of motion bias in VO/SLAM.
  
As one limitation of these results, features extracted with different filtering parameters and image resolutions introduce a scaling factor between the residual covariances and perspective deformation. So far, we extracted features at the original image resolution. The ORB implementation \cite{6126544} operates at discrete scale levels $s$ since it performs the same operations at different image resolutions. Figure \ref{fig:orb} shows the dependency of the residual covariance with the perspective deformation for each of these resolutions. For our model to be used at different scales, we approximate this effect by scaling our perspective deformation covariance, where $s$ and $s_r$ stands for the resolution factor of the reference and projected image respectively.   

Finally, we repeat the validation with real data the same manner as in Section \ref{subsec:photoPatches}. Figure \ref{fig:photoCal_fr} and Table \ref{table:geoCov_realData} show the results of the validation.

\begin{table}[h]
\centering
\begin{tabular}{c c c c c c c} 
\multicolumn{1}{c}{camera} & $\sigma_p$ & $\sigma^2_t$ & \multicolumn{1}{c}{$\sigma^2_c$}& $\varepsilon^2 \in$& $\mathtt{R}^2$ & $\frac{\sigma^2_t}{\sigma^2_c}$\\ [0.5ex] 
\hline
fr1       & 0.59 & 0.66 & 0.48 & 0.6-1.7 & 0.97 & 1.38\\ 
fr2       & 0.42 & 0.35 & 0.15 & 0.5-2.1 & 0.96 & 2.33\\
fr3       & 0.43 & 0.37 & 0.22 & 0.6-1.6 & 0.97 & 1.68\\
realSense & 0.56 & 1.88 & 1.42 & 0.8-1.2& 0.87 & 1.32\\
\end{tabular}
\caption{\textbf{Model validation for ORB features with real data}. The table shows the parameters of Equation \eqref{eq:geoCovSplit} estimated for the cameras of the RGBD-TUM dataset (fr) and a realsense D435i depth camera. Note how the visual covariance of ORB grows more in traction than in compression.}
\label{table:geoCov_realData}
\end{table}

\section{EXPERIMENTS} \label{sec:experiments}

The validation analysis in Section \ref{sec:modelVal} showed the relation between residual covariances and perspective deformation. In this section we demonstrate its applicability in state-of-the-art pipelines. Specifically, we evaluate the accuracy improvement in the photometric Bundle Adjustment (BA) of \cite{fontan2020information} and in the feature-based BA of ORB-SLAM \cite{mur2017orb}. 

For our evaluation we use the public TUM RGB-D benchmark \cite{sturm12iros}, that contains several indoor sequences captured with a RGB-D camera annotated with ground truth camera poses. Specifically, we use all static sequences except those beyond the range of the sensor. All the experiments were run on a standard laptop with an Intel Core i7-7500U CPU at 2.70 GHz and 8 GB of RAM for which the overhead caused by our model was less than 2\% of the total cost.

    \subsection{Information Metrics}  \label{subsec:information}

\begin{figure}[t]
\centering
\includegraphics[width=0.4\textwidth]{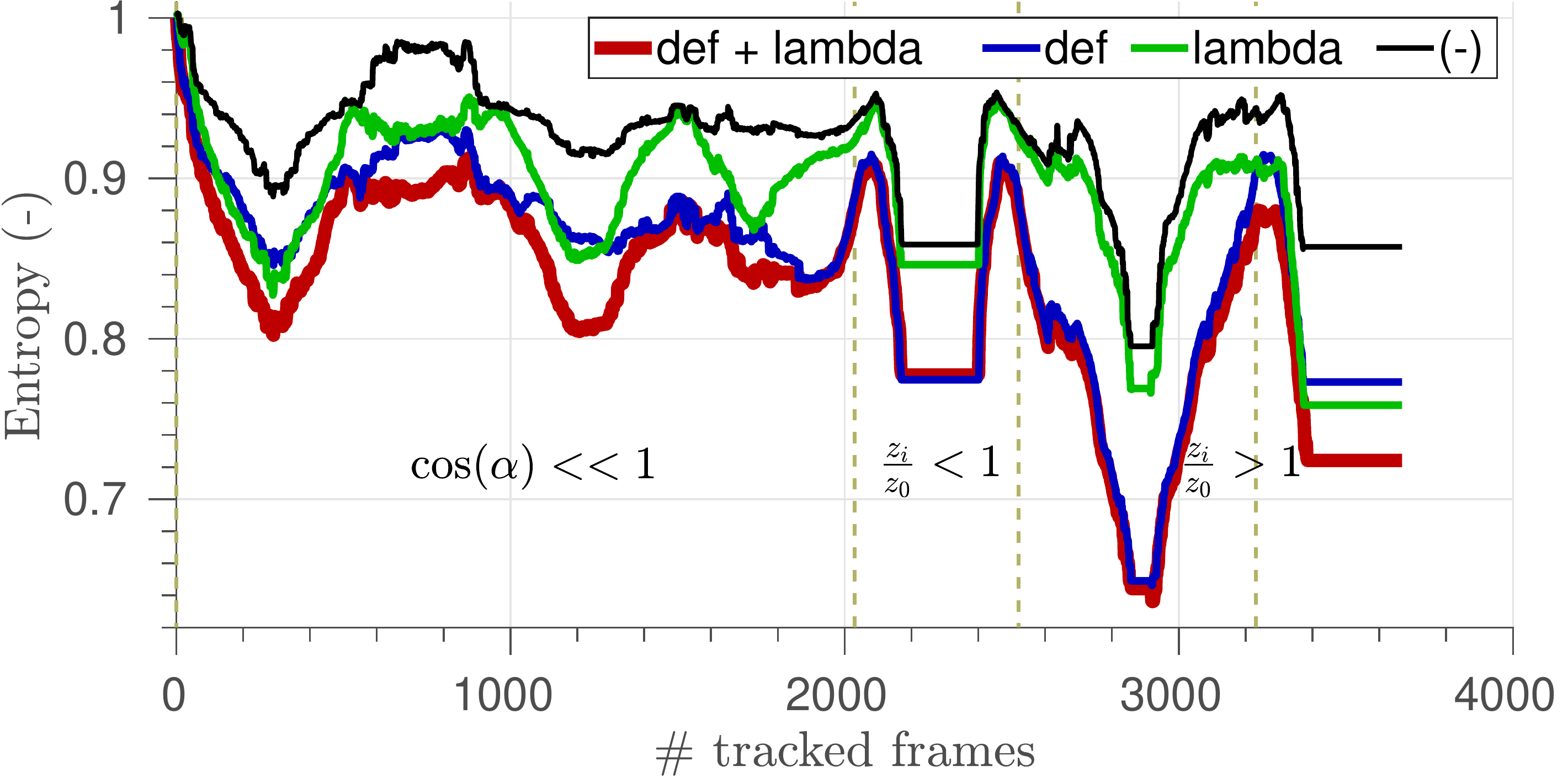}
\caption{Tracking entropy $H (bits)$. Big parallax ($\cos(\alpha) << 1$). Approximation ($\frac{z_i}{z_0} < 1$). Distancing ($\frac{z_i}{z_0} > 1$). Model complete (Def + lambda), just deformation covariance (def), just depth covariance (lambda), constant covariance (-).}
      \label{fig:pose entropy}
\end{figure}

As we anticipated, deriving the differential entropy of the camera pose $H(x) = -\frac{1}{2}\log((2\pi e)^k|\boldsymbol{\Sigma}_x|)$ from isotropic Gaussian residuals may lead to inconsistencies (see Figure \ref{fig:entropyInconsistencies}). Figure \ref{fig:pose entropy} show the effect of different additions to the covariance (Equation \eqref{eq:geoTerm}) in the pose estimation. We create a map from the very first RGB-D frame of sequence \textit{fr2 xyz} \cite{sturm12iros} and compute the available information to track each subsequent frame with respect to this initial map.

Figure \ref{fig:pose entropy} conveys at one glance the variation of the geometric covariance due to the propagation of depth uncertainty (Equation \eqref{eq:depthCov}) and due to perspective deformation (Equation \eqref{eq:geoCovSplit}). Note that, for motions producing big parallax ($\cos(\alpha) << 1$) the depth covariance is dominant. On the other hand, approximations ($\frac{z_i}{z_0} < 1$) or distancing motion ($\frac{z_i}{z_0} > 1$) produce strong perspective deformations. Our covariance model bridges the gap between visual errors and meaningful entropy values of the state.

    \subsection{Photometric odometry}  \label{subsec:idnav}

\textbf{Photometric BA.} ID-RGBDO \cite{fontan2020information} is a RGB-D direct odometry that uses information metrics for informative point selection and keyframe creation. ID-RGBDO performs BA over cameras and points in a sliding window. We implement at the end of each run a photometric global BA over all keyframes and points used along the sequence. We run the BA iteratively over this seed modifying poses and points with small Gaussian noise to observe the error distribution. We evaluate two models for the residual covariance: an isotropic Gaussian one and ours, based on deformations. Table \ref{table:idnav} collects the results in a selection of sequences of the TUM RGB-D dataset \cite{sturm12iros}. Specifically, we use all static sequences where the accuracy of the resulting trajectory is enough to guarantee that photometric BA converges, taking into account the smaller baseline for direct methods to converge. Note that our deformation model consistently leads to smaller trajectory errors.


\begin{table}[t]
\centering
\begin{tabular}{l c c | l c c } 
 Sequence & \cite{fontan2020information} & ours & Sequence &\cite{fontan2020information} & ours \\ [0.5ex] 
 \hline
 fr1. xyz &\textbf{1.55} &1.62      &  fr3. tex. str. far  & 1.56 & \textbf{1.45} \\ 
 fr1. rpy & 7.30 & \textbf{6.20}      &  fr3. tex. str. near & 1.89 & \textbf{1.78}\\  
 fr2. xyz & 0.90 & \textbf{0.81}     & fr3. tex. nstr. near & 3.89 & \textbf{3.52} \\ 
 fr2. rpy & 0.72 & \textbf{0.63}     & fr3. tex. str. far. v. & \textbf{1.22} & 1.98\\ 
 fr2. desk & 2.15 & \textbf{1.88}    & fr3. tex. str. near. v. & 4.50 & \textbf{2.36} \\ 
 fr2. dishes & 7.07 & \textbf{5.02}  & fr3. tex. nstr. near v. & 6.56 & \textbf{2.92}\\ 
 fr3. long office  & 3.11 & \textbf{2.65}  & fr3. long office v. & 2.95 & \textbf{2.26} \\ 

\end{tabular}
\caption{ATE (cm) for photometric BA in different sequences of TUM RGB-D. For each pair, the left one is the baseline with isotropic noise and the right one with our deformation model. Ours outperforms the baseline in 12/14 sequences, with an average ATE reduction of 12.6\% and a maximum reduction of 55.5\% in \textit{fr3. tex. nstr. near v.}.}
\label{table:idnav}
\end{table}

    \subsection{Feature-based SLAM}  \label{subsec:orbslam}

\textbf{Feature-based BA.} ORB-SLAM2 \cite{mur2017orb} is a feature-based SLAM system for monocular, stereo and RGB-D cameras. It includes some capabilities like map reuse, loop closing and relocalization. We run ORB-SLAM2 \rev {(where loop closure was deactivated from the original implementation in \cite{mur2017orb})} in different sequences and we apply a global BA at the end of each sequence over all the map points and all the keyframes poses. We modify this map by adding small Gaussian noise, in order to show variability in different runs. We then evaluate in different sequences two configurations for the global BA: with and without our deformation model. Table \ref{table:orbslam} shows the absolute trajectory error (ATE) of both configurations. Compared to the ATE of photometric BA (Table \ref{table:idnav}), notice two things. First, the tighter distribution of errors, confirming the better convergence of feature-based methods. And second, a smaller improvement, due to a higher degree of maturity of these methods and the complexity of modeling accurately the effect of the feature processing.

\begin{table}[t]
\centering
\begin{tabular}{l c c | l c c} 
 Sequence & \cite{mur2017orb} & ours & Sequence &\cite{mur2017orb} & ours \\ [0.5ex] 
 \hline
 fr1. xyz & 1.34 &\textbf{1.13}      &  fr3. tex. str. far  & 1.08 & \textbf{0.90} \\ 
 fr1. rpy & 3.17 & \textbf{3.09}     &  fr3. tex. str. near & 2.12 & \textbf{1.96}\\  
 fr2. xyz & 0.54 & 0.54     & fr3. tex. nstr. near & 1.37 & \textbf{1.21} \\ 
 fr2. rpy & 0.37 & \textbf{0.35}     & fr3. tex. str. far. v. & 1.12 & \textbf{1.04} \\ 
 fr2. desk & 4.15 & \textbf{3.99}    & fr3. tex. str. near. v. & 1.35 & \textbf{1.10} \\ 
 fr2. dishes & 4.67 & \textbf{4.47}  &  fr3. tex. nstr. near v. & 1.56 & \textbf{1.51}\\ 
 fr3. long office  & 2.39 & \textbf{2.33}  & fr3. long office v. & 2.34 & \textbf{2.06} \\ 

\end{tabular}
\caption{ATE (cm) for feature-based BA in different sequences of TUM RGB-D. For each pair, the left one is the baseline with isotropic noise and the right one with our deformation model. Ours outperforms the baseline in 13/14 sequences, with an average ATE reduction of 9.7\% and a maximum reduction of 22.7\% in \textit{fr3. tex. str. near. v.}.}
\label{table:orbslam}
\end{table}

\section{CONCLUSIONS AND FUTURE WORK}

In this paper we have derived for the first time a general model for the perspective deformation of 2-dimensional image patches and, based on that, we have particularized the relation of this deformation with feature-based and photometric residuals. We have validated the goodness of fit of the model in both synthetic and real data,  and we have shown experimentally that including perspective deformation into residual covariances improves the accuracy of direct and feature-based odometry and SLAM at a negligible computational cost and with minimal integration effort. Up to our knowledge, this is the first time that perspective deformation is explicitly modeled and applied to odometry and SLAM. We also show how to obtain more meaningful information metrics by modelling the covariances of the perspective deformation. Our evaluation focuses on global BA; since it is not coupled with other real-time parts of the pipelines (\emph{e.g.}, keyframe creation) and hence removes other factors from the evaluation. For future work we plan a tighter integration in a full tracking and mapping pipeline that relies on information metrics. 


\balance








{\small
\bibliographystyle{IEEEtran}
\bibliography{egbib}
}

\end{document}